\documentclass[10pt,journal,compsoc,twocolumn]{IEEEtran}

\usepackage{booktabs}
\usepackage{multirow}
\usepackage{flushend}
\usepackage{subfigure}
\usepackage{hyperref}
\usepackage{dsfont,amsfonts,amssymb,amsmath,color}
\usepackage{latexsym}
\usepackage{amsmath}
\usepackage{amsfonts}
\usepackage{algorithm}
\usepackage[noend]{algpseudocode}
\usepackage{graphicx}
\usepackage{color}
\usepackage{array}
\usepackage{algorithm}  
\usepackage{algpseudocode}
\usepackage{amsmath}  
\usepackage{makecell}
\usepackage{cases}
\usepackage{color}
\usepackage{amsthm}

\usepackage{soul}
\usepackage{tabularx}
\usepackage{makecell}
\usepackage{mwe}
\usepackage{fixltx2e}

\usepackage[colorinlistoftodos]{todonotes}
\usepackage{enumitem}
\usepackage{multirow}
\usepackage{multicol}
\usepackage{graphicx}
\usepackage{subfigure}
\usepackage{bm}
\usepackage{colortbl}
\usepackage{xcolor}
\usepackage[english]{babel}
\usepackage{xspace}
\usepackage{multirow}
\usepackage{rotating}

\usepackage{fancyhdr}
\usepackage[english]{babel}

\definecolor{Gray}{gray}{0.95}

\newcommand\SystemName{\textsc{Luce}\xspace}
\newcommand\HINName{\textsc{Hin}\xspace}
\newcommand\HIN{\textsc{Hin}\xspace}

\newcommand{\mypara}[1]{{\vspace{1mm}\noindent\textbf{#1}}}

\makeatletter
\def\hlinewd#1{%
  \noalign{\ifnum0=`}\fi\hrule \@height #1 \futurelet
   \reserved@a\@xhline}
\makeatother

\begin{document}

\title{Lifelong Property Price Prediction: A Case Study for the Toronto Real Estate Market}
%Higher-order and Attribute Enhancement Heterogeneous Graph Neural Networks

\author{~Hao Peng,
        ~Jianxin Li, ~\IEEEmembership{Member,~IEEE},
        ~Zheng Wang,
        ~Renyu Yang,~\IEEEmembership{Member,~IEEE},
        ~Mingzhe Liu,
        ~Mingming Zhang,
        ~Philip S. Yu, ~\IEEEmembership{Fellow,~IEEE},
        ~and Lifang He, ~\IEEEmembership{Member,~IEEE}
        % <-this % stops a space
\IEEEcompsocitemizethanks{
\IEEEcompsocthanksitem Hao Peng, Jianxin Li and Mingzhe Liu are with Beijing Advanced Innovation Center for Big Data and Brain Computing, Beihang University, Beijing 100083, China. E-mail: \{penghao, lijx\}@act.buaa.edu.cn, liumz@buaa.edu.cn.
\IEEEcompsocthanksitem Zheng Wang and Renyu Yang are with the School of Computing, University of Leeds, Leeds LS2 9JT, UK. E-mail: \{z.wang5, r.yang1\}@leeds.ac.uk.
\IEEEcompsocthanksitem Mingming Zhang is with the UrBrain Technology, Canada. E-mail: zmm021@gmail.com.
\IEEEcompsocthanksitem  Philip S. Yu is with the Department of Computer Science, University of Illinois at Chicago, Chicago 60607, USA. E-mail: psyu@uic.edu.
\IEEEcompsocthanksitem Lifang He is with the Department of Computer Science and Engineering, Lehigh University, Bethlehem, PA 18015 USA. E-mail: lih319@lehigh.edu.
}
\thanks{Manuscript received August 1th, 2020. (Corresponding author: Jianxin Li.)}
}

\markboth{}%
{Shell \MakeLowercase{\textit{et al.}}: Bare Demo of IEEEtran.cls for Computer Society Journals}

%==================================================================================
\IEEEtitleabstractindextext{%
\begin{abstract}
We present \SystemName, the first life-long predictive model for automated property valuation. 
\SystemName addresses two critical issues of property valuation: the lack of recent sold prices and the sparsity of house data. 
It is designed to operate on a limited volume of recent house transaction data. 
As a departure from prior work, \SystemName organizes the house data in a heterogeneous information network (HIN) where graph nodes are house entities and attributes that are important for house price valuation. 
We employ a Graph Convolutional Network (GCN) to extract the spatial information from the HIN for house-related data like geographical locations, and then use a Long Short Term Memory (LSTM) network to model the temporal dependencies for house transaction data over time. 
Unlike prior work, \SystemName can make effective use of the limited house transactions data in the past few months to update valuation information for all house entities within the HIN. 
By providing a complete and up-to-date house valuation dataset, \SystemName thus massively simplifies the downstream valuation task for the targeting properties. 
We demonstrate the benefit of \SystemName by applying it to large, real-life datasets obtained from the Toronto real estate market.
Extensive experimental results show that \SystemName not only significantly outperforms prior property valuation methods but also often reaches and sometimes exceeds the valuation accuracy given by independent experts when using the actual realization price as the ground truth.
\end{abstract}

\begin{IEEEkeywords}
Heterogeneous information network, graph neural network, LSTM, lifelong Learning, house price prediction.
\end{IEEEkeywords}}

%==========================================================================================

\maketitle

\IEEEdisplaynontitleabstractindextext

\IEEEpeerreviewmaketitle

\IEEEraisesectionheading{\section{Introduction}\label{sec:intro}}

For many families, a house is their most valuable asset.
Accurate and up-to-date house\footnote{In this work, a house is referred to as different types of residential properties, including the traditional house and apartments (or flats).} valuation is vital for various real-estate stakeholders such as homeowners, buyers, mortgage lenders, agents, etc.
House price estimation is traditionally performed by a real estate appraisal based on expert knowledge of target property, surrounding areas and historical data~\cite{bourassa2007spatial}, though at a very coarse granularity.
Substantial efforts -- most notably regression-based methods~\cite{basu1998analysis,park2015using,stevenson2004new} -- have been devoted to automate the house valuation by primarily examining the relationship between the house price and a range of quantified features like the property size, interior decoration, the number of bedrooms and facilities, the distance to a school catchment, etc.

Unfortunately, existing approaches for property valuation are inadequate in tackling two fundamental issues manifested by real-life property markets: data \textit{freshness} and \textit{sparsity}.
The key challenge here is that house transaction data are rarely up-to-date and inherently sparse - there is typically a gap of years between two transactions of a property and only a small number of houses are on the market for any given time.
For example, our analysis on the residential property transaction data of the Toronto Region in Canada between 2000 and 2019~\cite{datasource} shows that two consecutive transactions of a house typically spans over decades and only 0.1\% to 0.5\% of the residential properties within an administrative district (known as a neighborhood or community)\footnote{We consider
140 neighbourhoods (also often referred to as communities in the Canadian real estate market) officially recognized by the City of Toronto.} were traded within 12-month time frame.
Moreover, the small number of freshly traded houses spread across a large demographical area across thousands of households, making it difficult to effectively model and reason about the relationships between traded houses.
On top of that, transaction data before 2000 were often not in a digital form, which further reduces the availability of house transaction data.
The lack of current house transaction data implicates much of the pricing information that prior approaches rely upon cannot accurately reflect the market values of the target houses.
Given a complex and dynamic real estate market, the discontinuity and sparsity of house transactions make it extremely intricate to build an accurate predictor for house valuation.

To address the above limitations, we present \SystemName\footnote{\SystemName = \textbf{L}ifelong ho\textbf{u}se pri\textbf{c}epr\textbf{e}diction.}, a novel learning framework for lifelong house price prediction.
\SystemName is designed to work on a limited set of current house transaction data.
Our key insight is to use the most recent house transactions to estimate the value of all other properties of the target region (e.g., a metropolitan city).
By periodically updating and estimating the house information, \SystemName offers a life-long learning framework to estimate the current values for all houses with a metropolitan area.
By so doing, \SystemName enables the downstream house valuation model to utilize a significantly more substantial amount of house transaction data across many properties than what are available to prior methods.
The completeness and up-to-date house information provided by \SystemName thus enables one to build an accurate downstream valuation model using standard machine learning techniques.

Translating our high-level idea to build a practical system is, however, non-trivial.
Since \SystemName has to rely on a small number (i.e., data sparsity) of recent house transaction data across properties spreading across a large geographical area, it is important to make best use of all available data.
However, doing so is challenging because houses are distributed over a large geographical region with many attributes that can affect the house valuation.
To this end, we adopt the heterogeneous information network (HIN) to model the relationships - such as the location, facility, or floorplan - between houses entities.
We then employ a Graph Convolutional Network (GCN) to learn the house data representation, which is fed into a property valuation model built upon a standard multilayer perceptron (MLP) model.
Instead of directly performing learning on the entire, large HIN - which would be hard to generalize - we break down transactions into slices according to the geographical region of the traded house and when the transaction was taken place (on a monthly scale in this work).
This allows us to partition a large HIN into smaller sub-graphs so that the learning of house data representation can be performed on smaller graphs in parallel.
We go further by feeding the graph embeddings to a Long Short Term Memory (LSTM) network to improve the learned house representation by learning the temporal dependence of house data over time.

To address the discontinuity in house transaction data, we use the GCN-LSTM unit to perform house valuation of all house entities in the HIN over the last few months and then use the prediction and transaction history to estimate the price of the target house for the current month.
We show that this lifelong learning framework can be achieved by simply stacking up a sequence of GCN-LSTM learning units.
To overcome the gradient vanishing issue when performing learning over a long sequence of network layers, we introduce a sliding recursive parameter updating strategy to navigate the depth of gradient back-propagation and employ reinforcement learning to automate the parameter settings in the loss function calibration.
Our evaluation shows that this approach is simple to implement but yields good prediction performance.

We evaluate \SystemName by applying it to a real-world dataset collected from the Toronto real estate market. 
We compare \SystemName against 4 state-of-the-art automated house valuation methods and the valuation given by independent experts. 
When using the realization price as the ground truth, \SystemName outperforms all prior methods and often expert valuations.

This paper makes the following contributions. 
It is the first to:
\begin{itemize} [leftmargin=14pt]
    \item adopt an \HIN to model the house transaction data (~\S\ref{sec:problem-definition});
    \item propose a novel lifelong learning framework to perform property valuation (~\S\ref{sec:model} and
        ~\S\ref{sec:lifelong-struc});
    \item outperform prior automated house valuation methods and even expert property valuation on real-life datasets (~\S\ref{sec:eval}).
\end{itemize}

To enable replication and foster research we make
our \SystemName publicly available at: \texttt{\url{https://github.com/RingBDStack/LUCE}}.

\section{Background and Overview}\label{sec:problem-definition}

\subsection{Problem Scope and Motivation}

\mypara{Problem definition.} 
This work focuses on residential property valuation (price prediction). 
The prediction employs property specific information and transaction records to automatically estimate future property prices.
Our work addresses two primary research challenges facing the house valuation -- \textit{data sparsity} and \textit{data freshness} issues that manifest \textit{spatially} and \textit{temporally}. 
This is because only a small portion of houses is traded annually, while owner changing does not frequently manifest for most houses, resulting in a lack of up-to-date house transaction information temporally. 
Another challenge derives from the long-term learning wherein the vanishing gradients \cite{bengio1994learning} and catastrophic forgetting \cite{mccloskey1989catastrophic} effects of neural networks inevitably exhibit. 
Therefore, the learning model continuously learns on short-term dependencies but be lifelong so that we can replace the missing transaction information in the property network with the estimated values. 
Formally, we aim to learn a model that takes as input house features $\mathcal{X}$ and its previous sold price $\mathcal{Y}$, and house-related properties to predict the current valuation, $y_t$, of the target house $h$ at time $t$.

\mypara{Motivation.}  
Our work is motivated by the observation that prior regression-based work is insufficient to tackle data sparsity over time and stale transaction data cannot reflect realistic property prices. 
To illustrate this point, consider Table~\ref{table:rmse_comp} that gives the precision of different prediction approaches, Decision Tree Regression (DT), Support Vector Regression (SVR), and LSTM against the realization price for the residential property transaction data of Toronto between 2000 and 2019.
Regressors like SVR have a much higher Root Mean Square Error (RMSE) than the valuation given by human experts. 
Due to the intrinsic affinity among houses with similar terms of geographic location and floorplan design, we consider the price prediction as a regression problem based on nodes (houses) in a graph, where labels (i.e., house price) are only available for a small subset of nodes. 
We consider this problem as a life-long \textit{semi-supervised} learning based on graph embeddings.

%Dataset table
\begin{table}[t]
    \centering
    \scriptsize
    \caption{RMSE Comparison}
    \label{tab:dataset}
    \setlength{\tabcolsep}{2.5mm}
    \vspace{-1em}
    \begin{tabular}{rrrrr}
    \toprule
   \textbf{\#House Trans.} & \textbf{SVR} & \textbf{DT} & \textbf{LSTM-D} & \textbf{Appraiser-based} \\
    \midrule
    10,000 & 0.2677 & 0.2616  & 0.3055 & 0.1417  \\
    30,000 & 0.2792 & 0.2722  & 0.3047 & 0.1339  \\
    50,000 & 0.2885 & 0.2889  & 0.3076 & 0.1331  \\
    70,000 & 0.2987 & 0.2902  & 0.3021 & 0.1431  \\
    90,000 & 0.3011 & 0.2994  & 0.3015 & 0.1378  \\
    \bottomrule
    \end{tabular}
    \label{table:rmse_comp}
\end{table}

\begin{table}[t]
%\vspace{-0.5em}
\caption{Facility features and their types}
\label{tab:facility}
\vspace{-1em}
\scriptsize
\begin{tabular}{lllll}
\toprule\small
\textbf{Category} & \textbf{Type} & \textbf{Abbr.}  & \textbf{Type} & \textbf{Abbr.}\\
\midrule
\rowcolor{Gray}               & Townhouse & TH  & Detach & DE \\
\rowcolor{Gray}               & Semi-detach & SDE & Duplex & DU \\
\rowcolor{Gray}               & Triplex & TR & Fourplex & FO\\
\rowcolor{Gray}               & Cottage & CO & Link & LI\\
\rowcolor{Gray} \multirow{-5}{*}{\textbf{Building type}} & RuralResid & RR & Other & OT\\
~ & Backsplit & BA & Bungalow & BU \\
~ & OneNHalfStorey & ONS & TwoNHalfStorey & TNS \\
~ & TwoStorey & TWS & ThreeStorey & THS \\
\multirow{-4}{*}{\textbf{Layout structure}} & Sidesplit & SS & Other & OT \\
\rowcolor{Gray}               & Attach & A & Builtin & B\\
\rowcolor{Gray} \multirow{-2}{*}{\textbf{Garage type}} & Carpor & C & Detach & D\\

~ & AlumSliding & AS & Brick & BR \\
~ & Concrete & CO & MetalNSide & MS \\
~ & Shingle & SH & Stone & ST \\
\multirow{-4}{*}{\textbf{Exterior wall}} & VinylSling & VS & Wood & WO \\

\rowcolor{Gray}               & AboveGround & AG & Indoor & ID\\
\rowcolor{Gray} \multirow{-2}{*}{\textbf{Pool type}} & Inground & IG & None & NO\\

~ & Electricity & EL & Gas & GA \\
\multirow{-2}{*}{\textbf{Heat source}} & Oil & OI & Other & OT \\

\rowcolor{Gray}               & Baseboard & BB & FanCoil & FC\\
\rowcolor{Gray}               & ForcedAir & FA & HeatPump & HP\\
\rowcolor{Gray} \multirow{-3}{*}{\textbf{Heat equipment}} & Radiant & RA & Water & WA\\

~ & Crawl Space & CS & Step Entrance & SE \\
~ & Full & FU & Half & HA \\
~ & Finished & FI & Unfinished & UF \\
\multirow{-4}{*}{\textbf{Basement}} & Part Finished & PF & None & NO \\
\bottomrule
\end{tabular}
\end{table}

\subsection {Data Landscape} 
\begin{figure}[t]
	\center
	\includegraphics[width=0.45\textwidth]{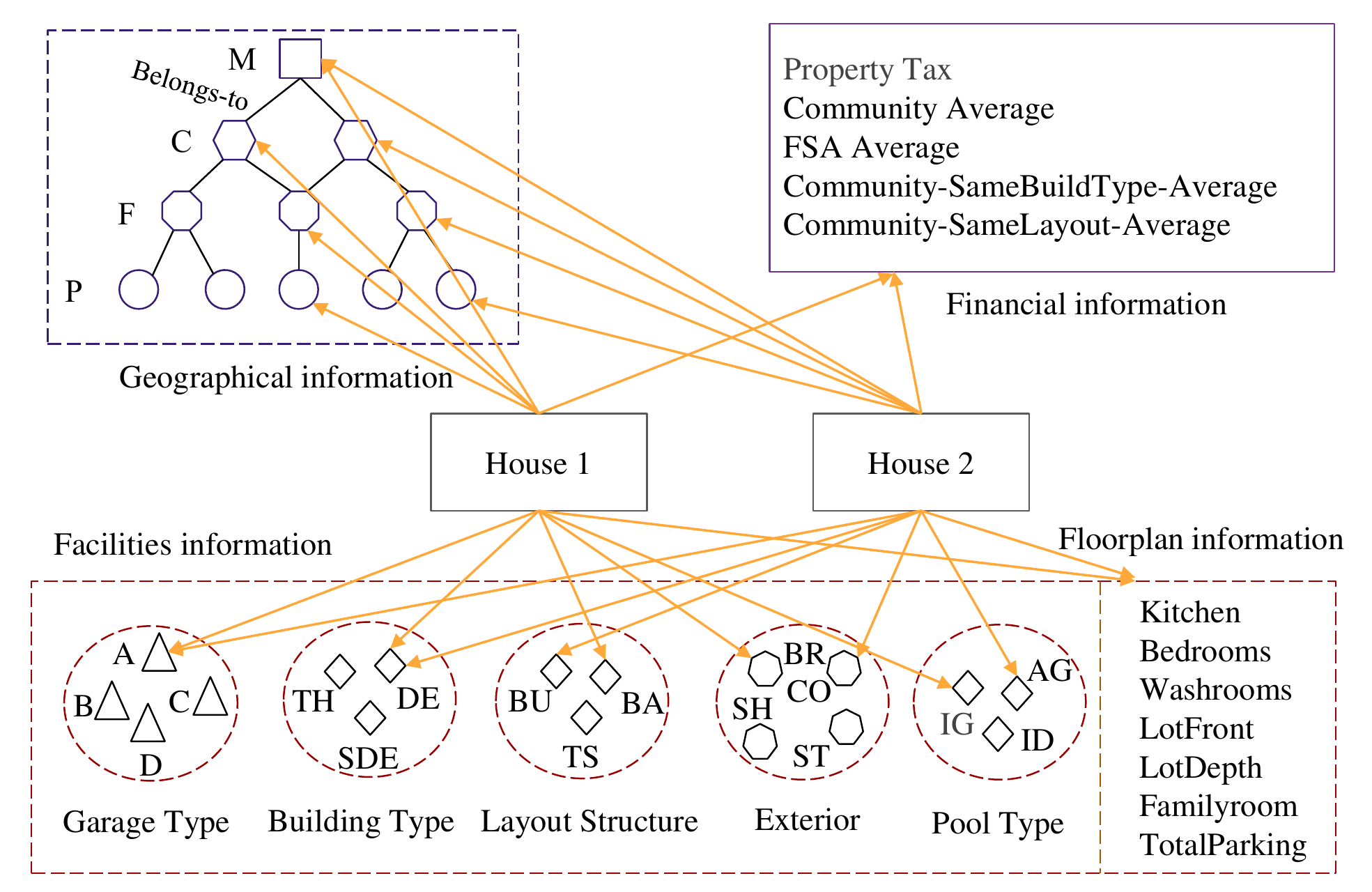}
	\vspace{-1em}
	\caption{House data and its HIN representation}
	\vspace{-1em}
	\label{fig:hhehone}
\end{figure}

While generally applicable, to have a realistic use case, our approach uses the house transaction data of the Region of Toronto between 2000 and 2019. 
This dataset is owned by the Toronto Real Estate Board~\cite{datasource}, an online information portal for real estate listings and services in the Greater Toronto area. 
The dataset consists of over a million transaction records of residential properties. 
Assumably, we can access the limited up-to-date information including the property size and floorplan because such information is often required to be supplied by the vendor to a real estate agent or lender. 
More information on the dataset can be found at \S\ref{sec:dataset}.
As depicted in Fig.~\ref{fig:hhehone}, we categorize the features into four distinct aspects:

% House space information---
%%%% note: the hierachical tree depicts the breakdown of a house location, rather than the collections of houses.  e.g., House 1 (M1, C1, F1, P1)
%% House space information---school information
%Compared with spatial information, school information is discrete and flat, and each school belongs to specific communities and forward sortation area.
%Each house can belong to specific spatial areas and school areas.
%We classify both spatial information and school information as house space information.
% internal information & supporting facility information

\textit{i) Geographical information:}
In our case study, we break down a valid house location by a series of address elements.
These elements can be formatted as a top-down hierarchical tree that encompasses different levels of geographical units.
As depicted in Fig~\ref {fig:hhehone}, the top of the tree is the largest and root geographical unit -- city-scale municipality (M) area.
Further down the tree, we sequentially define the geographical units as: the community (C) (i.e., neighbour in the Canadian system), the forward (F) sortation area (FSA)\footnote{A forward sortation area (FSA) is a geographical unit based on the first three characters in a Canadian postcode. 
All postcodes that start with the same three characters - for example, M4B - are together considered an FSA.}, and the postal (P).
Herein, the edge in the tree denotes the \emph{belongs-to} relationship between layers.
The geographical unit allows us to capture information like school catchments which are typically allocated through the geographical unit.

% The HIN figure

\textit{ii) Facilities information (enumerated):} 
Key indicators of the property valuation typically encompass supporting facilities (e.g., the type of garage, layout structure, etc.). 
Their types are \textit{enumerated} and summarized in Table~\ref{tab:facility}.

\textit{iii) Floorplan information (numerical):} the number of various rooms (bedroom, washroom, family room, kitchen, basement), the house area, width and depth of house land, and the number of stove, air conditioning and parking slots.

\textit{iv) Financial information (numerical):} 
We associate each property with its financial information (i.e., property tax) and the geographical unit information pertaining to the property. 
Specifically, the latter includes the average up-to-date property price of the community and the FSA, and the average price of properties of the same type within the community and the FSA.

\begin{figure}[t]
	\center
	\includegraphics[width=0.45\textwidth]{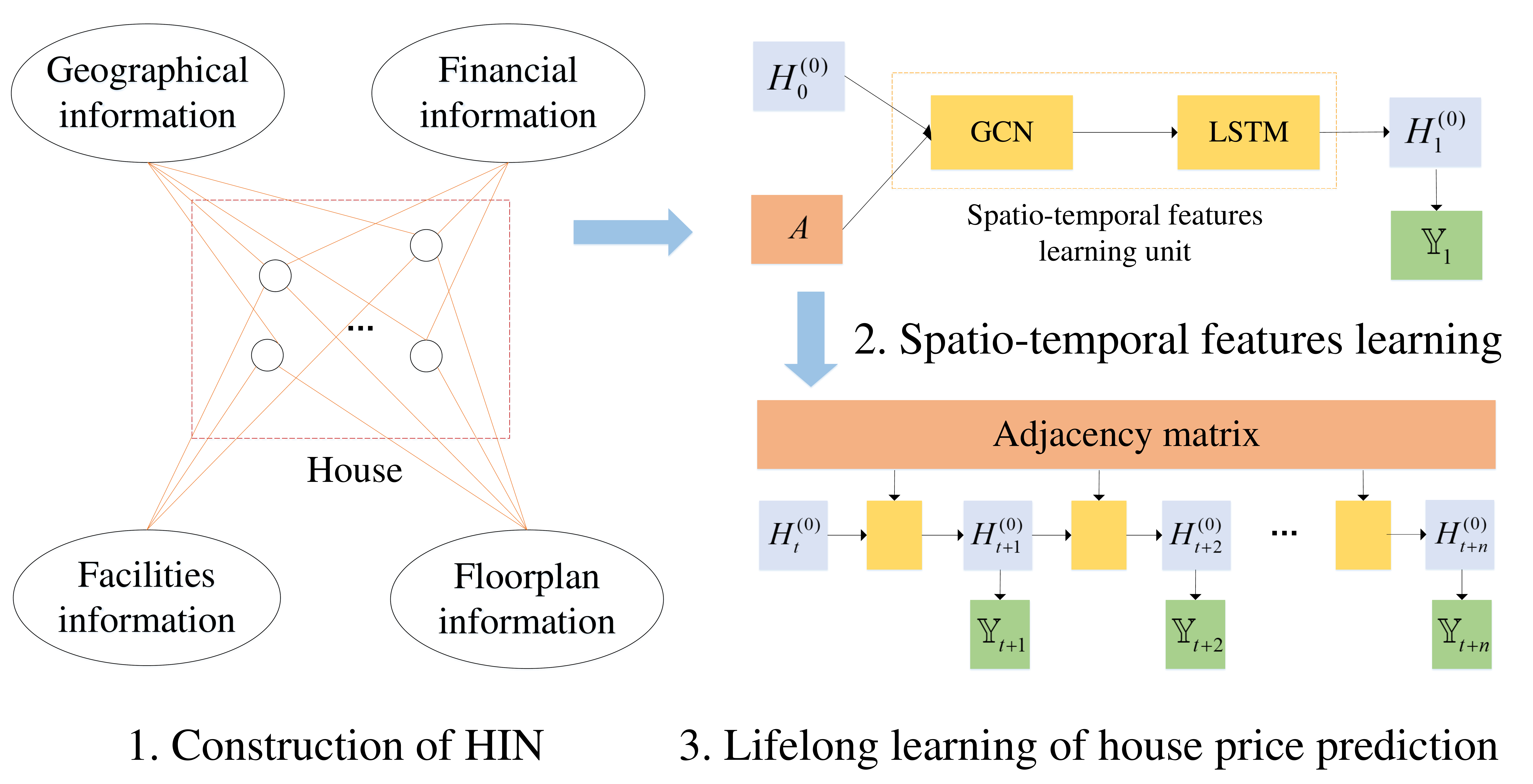}
	\vspace{-1em}
	\caption{Overview of \SystemName.}\label{fig:overview}
	\vspace{-1em}
\end{figure}

\subsection{\SystemName Solution}

\mypara{Overview.} The first innovation of our approach, as a departure from prior work, is to encode the information of house data as a structured, heterogeneous information network (HIN)\footnote{We refer to the same concept and definition about Heterogeneous Information Networks in previous work~\cite{sun2012mining,shi2016survey}.} wherein HIN nodes are different types of entities of houses and their characteristics, while edges represent different relationships between a pair of entities (e.g., a house \textit{belongs-to} a community/neighborhood, or a house has detached garage).

Since the problem of property price prediction is house entity oriented only, it is effective enough to deduce the information from a self-contained HIN to a homogeneous graph that can be directly absorbed by the graph convolutions.
In this context, the fundamental requirements of graph embedding for the HIN consists of three crucial elements -- obtaining graph structure and retaining node attributes (numerical features) and node label.
In fact, the graph structure reflects the \textit{structural connectivity} between two house entities, based on the affinity in terms of geographic proximity and pertaining facilities.
Meanwhile, numerical \textit{attributes} of an individual house regarding the detailed floor plan should be maintained as the main features while property price is regarded as node label.

At the core of \SystemName is a deep neural network that builds upon the GCN and LSTM.
The network learns the most appropriate embedding for the structural and numerical features captured by the HIN and uses the learned representation and known labels to perform the price regression.
Most notably, the semi-supervised GCN allows for feature learning for all houses through a limited number of available transaction sourcing from the sparse houses.
Herein, we use the native GCN as the representative instance -- due to its simplicity and general purpose use cases -- while any other latest semi-supervised graph neural network models~\cite{VeliGraph,wu2020comprehensive} can be easily used as a substitute for the GCN in our scenario of house feature learning.

% The meta-graph figure
\begin{figure*}[t]
    \center
    \subfigure[Meta-schema of \HINName.]{
        %\begin{minipage}[c]{0.49\textwidth}
        %\centering
		\includegraphics[width=0.44\textwidth]{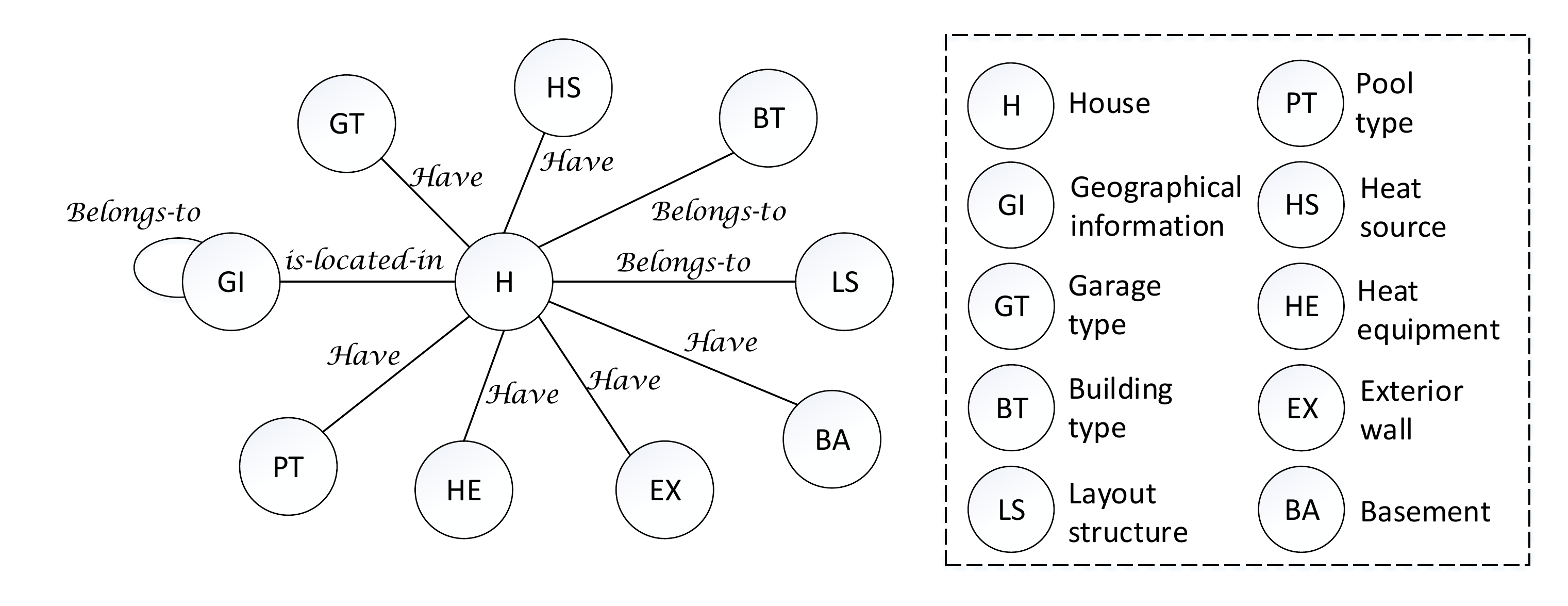}
		\setlength{\leftskip}{-20pt}
%		\vspace{-0.4em}
		\label{fig:metaschema}
		%\end{minipage}
	}
    \subfigure[Examples of meta-paths.]{
        %\begin{minipage}[c]{0.24\textwidth}
        %\centering
		\includegraphics[width=0.24\textwidth]{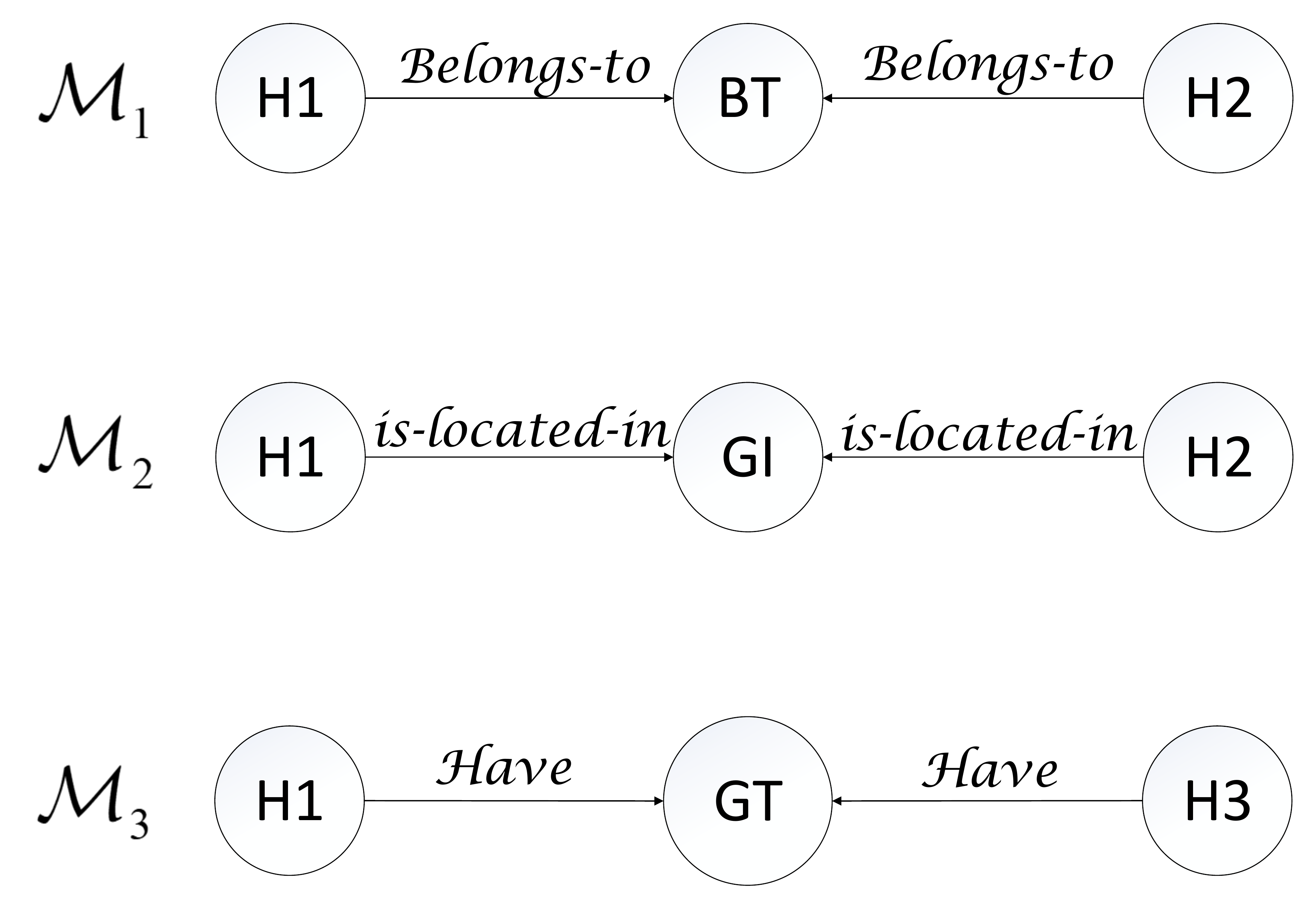}
		\setlength{\leftskip}{-20pt}
%		\vspace{-0.4em}
		\label{fig:metapath}
		%\end{minipage}
	}
	\subfigure[Examples of meta-graphs.]{
	    %\begin{minipage}[c]{0.25\textwidth}
        %\centering
		\includegraphics[width=0.24\textwidth]{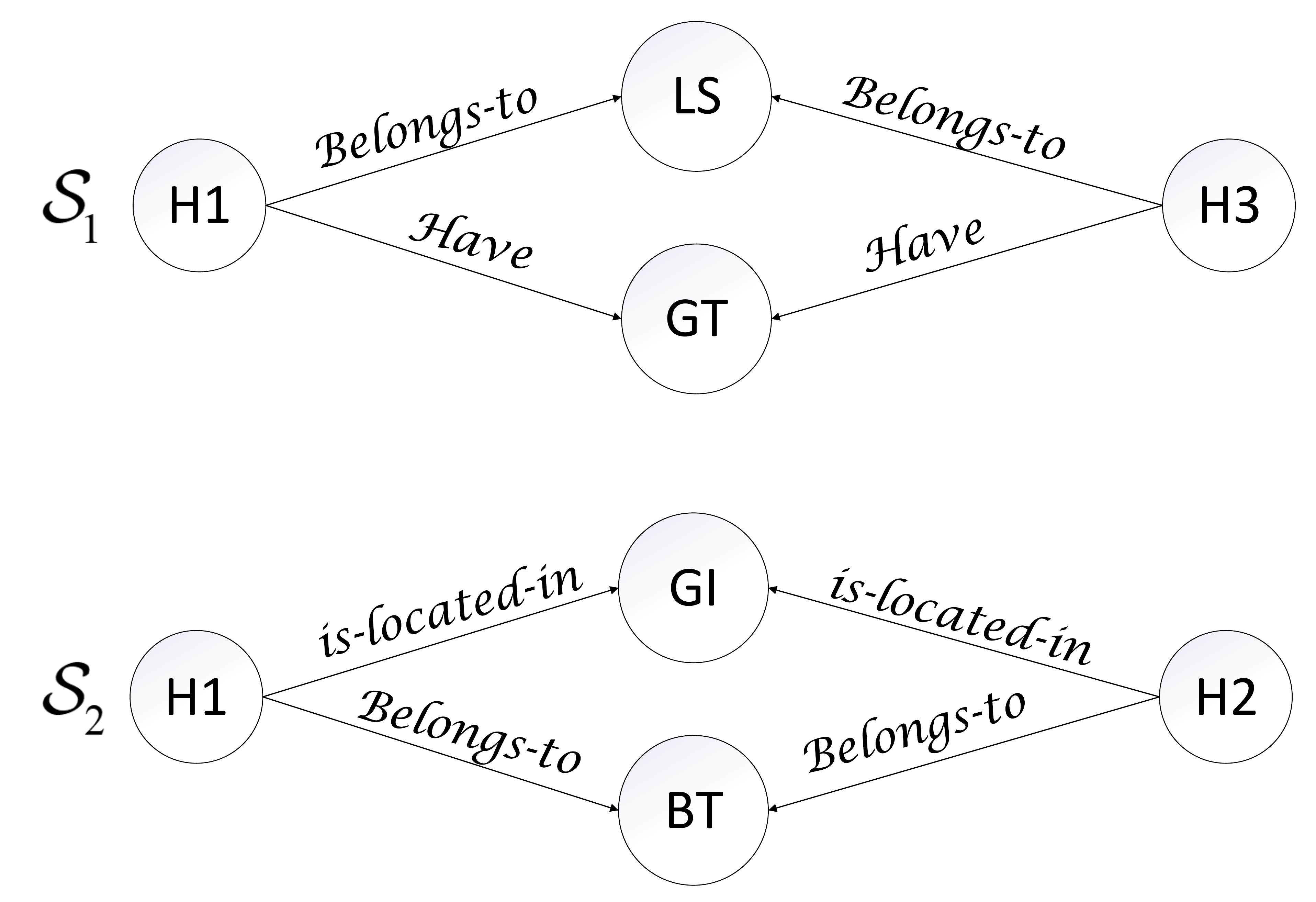}
		\setlength{\leftskip}{-20pt}
%		\vspace{-0.4em}
		\label{fig:metagraph}
		%\end{minipage}
	}
    \vspace{-1em}
	\caption{(a) Meta-schema denotes basic relationships among entities in \HINName; (b) a meta-path encodes a common relationship/feature shared by two entities; (c) a meta-graph encodes multiple relationships shared by a pair of entities.}\label{fig:meta}
    \vspace{-1em}
	\setlength{\leftskip}{-40pt}
\end{figure*}

\begin{table}[t]
\scriptsize
    \centering
    \caption{Notations}\label{tab:symbol}
    \vspace{-1em}
    \setlength{\tabcolsep}{1.6mm}
    \begin{tabular}{r|l}
    \toprule
   \textbf{Symbol} & \textbf{Definition} \\
    \midrule
    $\mathcal{G}$; $\mathcal{V}$; $\mathcal{E}$ & HIN and its node sets and edge sets\\
    $\mathcal{X}$ & House features set \\
    $\mathcal{Y}$ & Node label (house sold price) set\\
    $t$; $T$ & Time step number; Total time steps (months)\\
    $Sim(v_{i}, v_{j})$ & The similarity of house $v_{i}$ and house $v_{j}$\\
    $A$ & Adjacency matrix based on house similarity\\
    $\vec{\omega}$ & The weight of meta-path or meta-graph\\
    $\mathcal{M}'$ & The total number of meta-paths and meta-graphs\\
    $\mathbb{L}_t$ & Loss function of at time step $t$ \\
    $y_{t}$ & The sold-price set of house at time step $t$\\
    $H_{(i,t)}$ & The house embedding generated by $i$-th GCN at time step $t$\\
    $\mathbb{H}_{t}$ & The house embedding at time step $t$\\
    $\mathbb{Y}_{t}$ & The predicted prices of all houses at time step $t$\\
    $\mathcal{W}_{(i,t)}^{(l)}$ & The parameter matrix of $l$-th layer of $i$-th GCN at time step $t$\\
    $\theta_{(i, t)}$ & The parameter of LSTM of $i$-th subgraph at time step $t$\\
    \bottomrule
    \end{tabular}
\end{table}

It is worth noting that, unlike prior work that directly operates on the discrete and sparse time-series data across years, \SystemName splits available transactions into monthly slices and uses data within each month slice to train the GCN-LSTM continuum, the basic feature learning unit.
This is because the house price is observably stable between two consecutive months and we can constantly update the prediction model on a monthly basis.
Accordingly, the trained network manages to predict \textit{all} house prices in the coming months.

In addition, the contribution of different meta-paths and meta-graphs in the HIN can be learned.
In order to avoid out-of-memory, we can naturally divide the graph of houses into subgraphs, jointly connected by houses within the overlapping areas.
Correspondingly, we break down the basic GCN-LSTM training unit into independent and parallel GCN-LSTM instances, each of which is exploited for feature embedding in each subgraph.
Such parallelism will finally form an array of GCN-LSTM, thereby significantly speeding up the procedure of feature learning.

\mypara{Architecture and pipeline.}
Fig.~\ref{fig:overview} depicts the overall architecture of \SystemName.
From the constructed HIN, we firstly calculate the adjacency matrix -- the best option to reflect the proximity and the node connectivity in the graph -- and the attribute matrix to retain the residual numerical features of nodes.
Due to the intrinsic fact that GCNs merely operate on homogeneous graph and induce embedding vectors for nodes based on the properties of their neighborhoods, we compute the similarity between every pair of houses and store it with adjacency matrix, to underpin the heterogeneous graph convolution (\S\ref{sec:HHe-HoNe}).

We feed adjacency matrix $A$ -- generated by both meta-paths instances and meta-graphs instances based similarity measurement -- together with the house (or node) into a GCN and a graph LSTM.
The learned embedding vectors delivered by the GCN-LSTM continuum are therefore concatenated to form the holistic representation of the original \HIN at time $t$ (\S\ref{sec:GCNLSTM}).

To build a lifelong prediction framework capable of estimating house price monthly, we design a multitask learning scheme where GCN-LSTM units are unfolded multiple times in a pipeline. 
A network obtained at month $t$ can be inherited for predicting the house prices $\mathbb{Y}_{t}$ in the coming month $t+1$. 
Meanwhile, the prediction will also update the embedding of houses $\mathbb{H}_{t+1}$ at time $t+1$, which is used to train the follow-up GCN-LSTM units. 
We iterate this process until targeting the valuation of all houses at a certain month, $t+n$. 
To deal with the gradient vanishing problem, we introduce a sliding recursive strategy for parameter updating -- limiting the depth the gradient can back-propagate and recognizing the different impact of the embedding within each time on the loss function calibration. 
We use reinforcement learning to auto-learn the parameters involved in the calibration(\S\ref {sec:lifelong-struc}).
To aid discussion, Table \ref{tab:symbol} depicts the notations used throughout the paper.

\section{Temporal-Aware Network}\label{sec:model}
\subsection{Graph Construction from \HINName}\label{sec:HHe-HoNe}
\mypara{Calculating meta-path and meta-graph.} 
Meta-schema is a meta-level template that defines the relationship and type constraints of
nodes and edges in the HIN. 
As shown in Fig.~\ref{fig:metaschema}, we obtain a meta-schema that encodes all possible relationships between
the house entity and other types of entities. \textit{Meta-path} is a path that connects a pair of network nodes with a semantically meaningful relationship between nodes (exemplified in Fig.~\ref{fig:metapath}). We can enumerate all existing relationships
among each pair of house entities as the pre-defined meta-paths. In fact, a meta-path can be used to encode common features shared by two
houses, e.g., two houses belong to the same community. As a pair of houses could have an arbitrary number of meta-paths,
\textit{meta-graph}, in the form of directed acyclic graph (DAG), can be used as a template to capture the arbitrary but meaningful
combination of existing meta relationships between a pair of nodes. For instance, the meta-graphs described in Fig.~\ref{fig:metagraph}
define two templates -- houses have the same layout and garage type (above) and houses located in the same area have the same building type
(below).

%%\yry{field expertise is required to define the meaningful meta graph? how to provide the set of meta graphs? do we need to specify the number of meta paths included in a graph? or purely enumerate all possibilities?}

%For we use multiple meta graphs to encode different meta-path combinations, meta paths between two houses entities can be mapped onto more than one meta graph.

% We also represent the financial information as a vector of numerical attributes, and then concatenate financial, facilities and floorplan information as a united vector of numerical attributes as house (node) property.

% After extracting the above heterogeneous house property information, we build the HIN to model houses. For better understanding, as
%depicted in Figure~\ref{fig:metaschema}, we also give a meta-level (i.e., schema-level) description or meta-schema\footnote{We use the same
%concept of meta schema defined in previous work~\cite{Shi2017A} in our task.}, where meta-schema is a relational graph defining the type
%constraints of nodes and edges in the HIN. For example, Figure \ref{fig:hhehone} can be regarded as an instance of the meta-schema in
%Figure \ref{fig:metaschema}. One advantage of the HIN based house modeling is that meta-paths and meta-graphs defined over types and
%structures (introduce in Section~\ref{sec:similarity-measure}) can reflect semantically meaningful information about similarities, and thus
%can naturally provide explainable analysis and result for house similarity measurement.

\mypara{Retrieving structural information.} Meta-paths and meta-graphs over types and structures indicate semantic-explainable similarities
between two houses -- houses have more meta-path instances and meta-graph instances tend to have closer valuation. In addition, different
meta-graphs should be arguably differentiated when computing the similarity according to the differed semantic implications in meta-paths.
For example, a house within the same postal area with the same layout structure as house $x$ is far more likely to have similar valuation
compared against another house that has the same building and pool style as house $x$ but in a different community.

Following similar methodology presented in \cite{Peng2019Fine}, we compute the similarity of house valuation between two houses $h_i$ and $h_j$ as follows:
\begin{equation}
\label{eq:house_sim}
\footnotesize
\vspace{1em}
S(h_i,h_j) = \sum_{m=1}^{M'}\omega_{m}\frac{2\times Count_{C_m}(h_{i},h_{j})}{Count_{C_m}(h_{i},h_{i})+Count_{C_m}(h_{j},h_{j})},
\end{equation}
where $C$ denotes the collection of meta-paths and meta-graphs, and $Count_{C_m}(h_{i},h_{j})$ counts the number of $m$-th element (path/graph) $C_{m}$ between two house instances $h_i$ and $h_j$.
Similarly, $Count_{C_m}(h_{i},h_{i})$ and $Count_{C_m}(h_{j},h_{j})$ compute the number of meta-paths and meta-graphs instances between $h_i$ and $h_i$, and between $h_j$ and $h_j$, respectively.
The number of meta-graph instances is counted by the \emph{Hadamard product} between matrixes counted by sub-string meta-paths.
At the core of the similarity function, $S(h_i,h_j)$, is to normalize the importance of meta-paths and meta-graphs between $h_i$ and $h_j$ by applying different weights to different structural relationships.
$2\times Count_{C_m}(h_{i},h_{j})$, counts the number of shared meta-path instances and meta-graph instances between house instances $h_i$ and $h_j$ for computing the semantically overlapped information.
We multiply the number by two because the meta-paths and meta-graphs are bi-directional.
$Count_{C_m}(h_{i},h_{i})+Count_{C_m}(h_{j},h_{j})$ counts the total number of meta-paths and meta-graphs among the two house instances themselves.
Notably, we use a \textit{learnable} parameter vector $\vec{\omega} = [\omega_{1},\omega_{2},$ $\dots,\omega_{M'}]$ to denote weights of all meta-paths and meta-graphs.

We can then use the calculated similarity to indicate the connectivity between any pair of house instances.
Accordingly, we construct an $N \times N$ weighed adjacency matrix $A$ to store the semantic similarity among $N$ houses.

\mypara{Retrieving house attribute matrix through PCA.}
% For non-numeric house geographical units and facilities information, we use the concatenated one-hot representation.
There are several house-related numerical attributes. 
We use one-hot representation to encode each of these numerical attributes, after which
we concatenate them as a single vector of numerical values. 
An attribute vector is therefore associated with a house entity in the \HIN. 
We further apply principal component analysis (PCA) to reduce the dimensionality of the vector to $D$ (e.g., to $100$ elements). 
In this manner, we eventually form the house attribute matrix $X$, which is of shape $N\times D$.

%
% Next, we employ graph convolutional neural network and
%LSTM to learn spatio-temporal house (node) representation. With the help of similarity weighted adjacency matrix $A$, the embedding learned
%by the graph neural network can effectively overcome the problem of sparsity of reference houses.

% Spatio-temporal features learning unit figure
\begin{figure}[t]
	\center
	\includegraphics[width=0.5\textwidth]{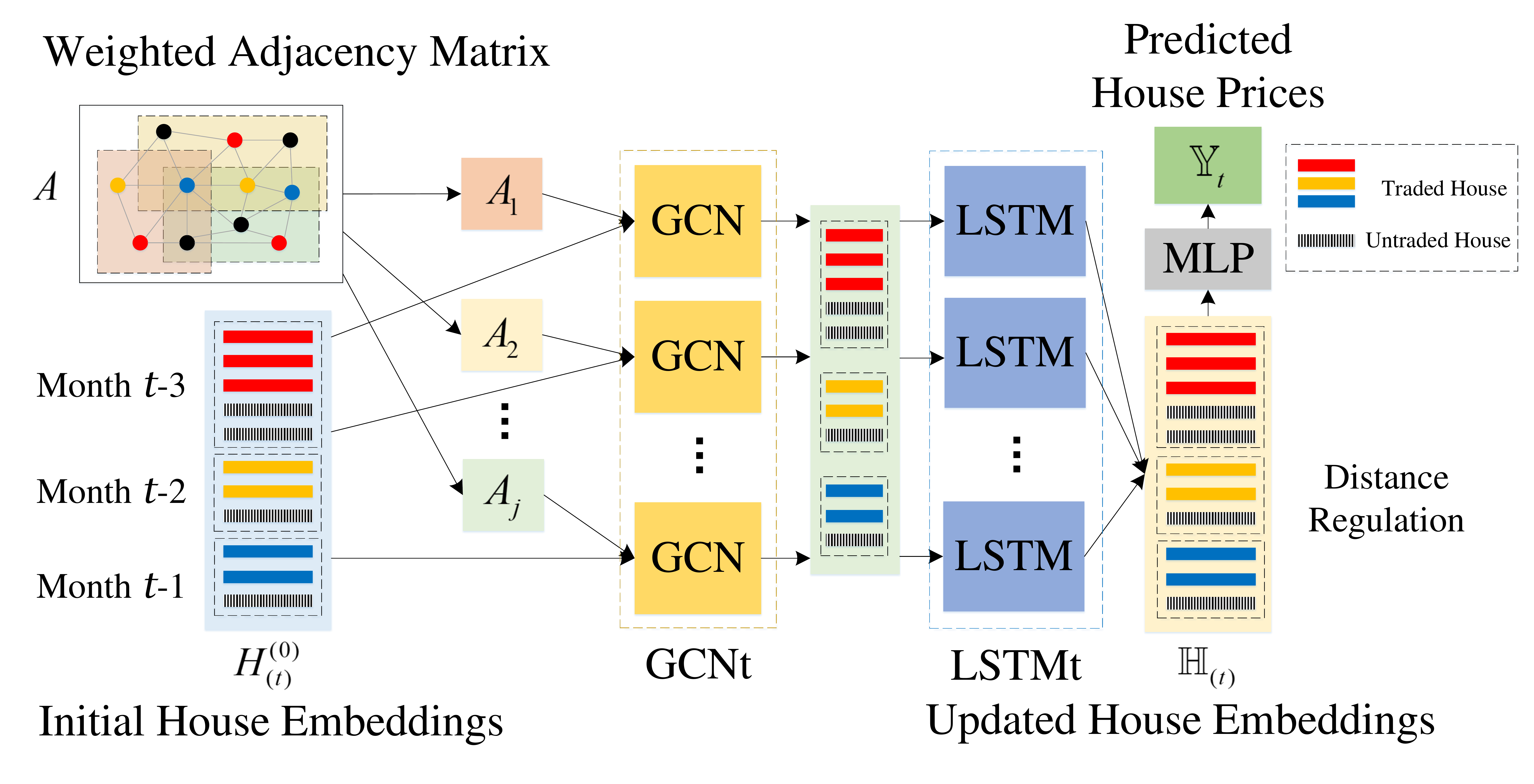}
	\vspace{-1em}
	\caption{Basic unit of feature learning} %% to match the unit in Fig.5}	}
	\vspace{-1em}
	\label{fig:glunit}
\end{figure}

% sub-model introduction
\subsection{Temporal-Aware Feature Learning}
\label{sec:GCNLSTM}
%We first propose a neural network unit with spatio-temporal prediction ability to overcome the sparsity of house transaction records and be able to deal with short-term discrete temporal features.
%We name this neural network combined graph convolutional network layer and LSTM layer as a spatio-temporal feature learning unit, and Figure \ref{fig:glunit} shows its structure.

Fig.~\ref{fig:glunit} illustrates the flowchart of \SystemName learning framework.
The trunk of \SystemName is to take as input the weighted adjacency matrix $A$ and the house attribute matrix $X$, and constantly update the house embedding $\mathbb{H}_{t}$ at the $t$-th month.
The house price $\mathbb{Y}_{t}$ can be then predicted based on the embedded features with known \textit{price label}.

% geo features --> non-numerical
% facility features --> non-numerical, enumerated types
% floorplan features  -> numerical
% financial features -> numerical

%% the number of facilities (stove, air conditioning,parking space? which category?

%% meta-path/graph only consider non-numerical features? --> adjacency matrix (graph structure) + still need representation (e.g., feature vector)
%% other numerical features --> attribute matrix, representation is achieved..
At the core of the graph embedding is learning the representation of house features.
We leverage GCN to aggregate the neighborhood information in $A$ when measuring the relationship between an entity pair.
Primarily exploiting the weighted adjacency matrix and attribute matrix, GCN learns and feeds the feature representation to a LSTM to involve temporal dependency within the \HINName and further calibrate the effectiveness of the graph embedding.
We introduce parallelism to accelerate the model training and reduce the memory overhead when processing a large graph. As houses have been intrinsically divided into adjacent geographic areas, we split the holistic graph into several overlapping subgraphs, and conduct feature learning for each subgraph in parallel. Specifically, we divide the weighted adjacency matrix $A$ into several overlapping subgraphs (see Fig.~\ref{fig:glunit}) and it can be formalized as:
\begin{align}
\label{eq:multi-graphs}
\footnotesize
    A = A_{1} \cup A_{2} \cdots \cup A_{j},
\end{align}
where $j$ is the total number of divided subgraphs.

\mypara{Feature learning.} For the $i$-th subgraph $A_i$, we employ a GCN model~\cite{KipfSemi} to learn the numerical feature embedding on a monthly basis, by formalizing a layer-wise propagation rule at the $t$-th month:
\begin{align}
\label{eq:GCN}
\footnotesize
   H_{(i, t)}^{(l)} = GCN(A_{i}, H_{(i, t)}^{(0)}, W_{(i, t)}^{(l)}),
\end{align}
where $H_{(i, t)}^{(0)}$ is the afferent feature matrix of the $i$-th subgraph, and $W_{(i, t)}^{(l)}$ is the parameter matrix of the $i$-th subgraph at $l$-th layer.
%\begin{align}
%\label{eq:GCN}
%\footnotesize
%    H_{t}^{(l+1)}=\sigma (\tilde{D}^{-\frac{1}{2}}\tilde{A}\tilde{D}^{-\frac{1}{2}}H_{t}^{(l)}W_{t}^{(l)}),
%\end{align}
%where $\tilde{A}=A_i+I$, $A_i$ is the $i$-th divided weighted house similarity adjacency matrix, and $I$ is the identity matrix.
%$\tilde{D}$ is the degree matrix of $\tilde{A}$, and $\tilde{D}_{ii} = \sum_{j}\tilde{A}_{ij}$.
%$W_{t}^{i}$ is the parameter matrix at $l$-th layer while $H_{t}^{(l)}$ is the feature matrix of the $l$-th layer.
$H_{(i, 1)}^{(0)}$ is set to be the initial house attribute of the $i$-th sub-graphs from $X$.
%$\sigma$ is an activation function such as Sigmoid or ReLU.
After training, we record the $l$-th layer embedding of the $i$-th subgraph at $t$-th month as $\mathcal{H}_{(i, t)} = H_{(i, t)}^{(l)}$.
In fact, the adjacency matrix $A_i$ contains the parameter $\vec{\omega}$ pertaining to each meta path/graph in the similarity measurement.
Notably, the parameter will be continuously and globally updated in the training procedure.

This semi-supervised GCN technology allows for learning embedded features of all houses using a limited number of available house transactions.
Meanwhile, it enables us to learn the contribution of different meta-paths and meta-graphs, which can facilitate to divide house instances into independent and parallel GCNs.
Still, it is intractable how to leverage house transactions at different time (e.g., different months) to calibrate the house embedding more precisely.
To tackle this, we will work on the temporal features at different time periods.

%For GAT, we utilize the graph attention layer with multi-head attention mechanism\cite{VeliGraph}. The input of this layer is the house embedding $h=\{\vec{h}_{1},\vec{h}_{2},\cdots,\vec{h}_{n}\}$, and the output is the updated house embedding $h'=\{\vec{h}'_{1},\vec{h}'_{2},\cdots,\vec{h}'_{n}\}$. The calculation method for $h’$ is:
%\begin{align}\label{eq:GAT1}
%    e_{ij}=\text{LeakyReLU}\left(\vec{a}^{T}\left[W\vec{h_{i}}||W\vec{h_{j}}\right]\right)
%\end{align}
%\begin{align}\label{eq:GAT2}
%    \alpha_{ij}=\text{softmax}_{j}(e_{ij})=\frac{\text{exp}(e_{ij})}{\sum_{k\in \mathcal{N}_{(i)}}\text{exp}(e_{ik})}
%\end{align}
%\begin{align}\label{eq:GAT3}
%    \vec{h}'_{i}=||^{K}_{k=1}\sigma\left(\sum_{j\in \mathcal{N}_{(i)}}\alpha_{ij}^{k}W^{k}\vec{h}_{j}\right)
%\end{align}

%Equation (\ref{eq:GAT1}) calculates the attention coefficient $e_{ij}$ of house $i$ and house $j$, where $W$ is the parameter matrix, $\vec{a}$ is the parameter vector, and $||$ represents the vector concatenation. Equation (\ref{eq:GAT2}) calculates the weight coefficient $\alpha_{ij}$ through the \textit{softmax} function, $\mathcal{N}_{(i)}$ is the neighbor node set of $i$. Equation (\ref{eq:GAT3}) is the multi-head attention mechanism, which uses the weight $\alpha_{ij}$ to aggregate the information of multiple neighbor nodes to represent the embedding of the house $i$.

\mypara{Temporal dependency.} House embedding obtained from the GCN cannot guarantee an up-to-date price information due to the ignorance of time difference in the price label.
We therefore add an additional LSTM layer to learn and update the valuation for each house.
Specifically, we feed the learned house embedding $\mathcal{H}_{(i, t)}$ of the $i$-th subgraph at $t$-th month as the input of LSTM units.
The output of the LSTM can be formalized as:
\begin{align}
\label{eq:LSTM}
\footnotesize
\mathbb{H}_{(i, t)} = LSTM(\mathcal{H}_{(i, t)},\theta_{(i, t)}), \quad H_{(i, t+1)}^{(0)} = \mathbb{H}_{(i, t)}
\end{align}
where $\theta_{(i, t)}$ means parameters in LSTM unit and the output of LSTM unit will be passed to next GCN unit as the initial house attribute.
Consequently, the feature embedding of houses are transformed into time series according to the transaction times, significantly alleviating the discontinuity of house transactions in the short run. We then concatenate all individual embeddings $\mathbb{H}_{(i,t)}$ into $\mathbb{H}_{(t)}$ before making fine-grained calibration. Hence, for the month $t$, by using the transaction prices of houses in the \HINName, we can train evolving house embeddings, which integrate both spatial and temporal features.
Eventually, we add a multi-layer perceptron (MLP) between the delivered embedding by LSTM and the price label to decode and thus predict any house prices $\mathbb{Y}_(t)$, at $t$: $\mathbb{Y}_t=MLP(\mathbb{H}_{(t)})$.

\mypara{Calibration through distance regulation.} We instantiate an independent GCN for each subgraph to obtain its own feature embedding in parallel. To coordinate those embedding results and form a holistic picture, we use \textit{distance regulation} to calibrate the embedding, ensuring the house across different subgraphs has close embedding scheme in different GCNs:
\begin{align}
\label{eq:regularization}
\footnotesize
\epsilon(P_t) = \sum_{\alpha=1}^{L}||\mathbb{H}_{(1,t)}(p_{\alpha})-\sum_{\beta=2}^{g(p_{\alpha})}\frac{1}{g(p_{\alpha})-1}\mathbb{H}_{(\beta,t)}(p_{\alpha})||,
\end{align}
where $\mathcal{P}$ denotes the set of those overlapping houses, i.e., $\mathcal{P}=\{p_{\alpha}, \alpha=1, \dots ,L\}$. 
$g(p_{\alpha})$ means the number of subgraphs that contain the $p_{\alpha}$ house, and $\mathbb{H}_{(\beta,t)}(p_{\alpha})$ refers to the embedding of the $p_{\alpha}$ house in the $\beta$-th GCN at the $t$-th month.

\mypara{Loss Function.} We use the following loss function to optimize model parameters:
\begin{align}
\label{eq:object}
\footnotesize
\mathcal{L}_{t}= \mathcal{R}_{t} + \epsilon(P_t),
\end{align}
The loss function $\mathcal{L}_{t}$ comprises two parts -- the accuracy of prediction -- the Root Mean Square Error (RMSE)  $\mathcal{R}_t$ -- and the unified embedding of overlapping houses among GCNs. The widely-used stochastic gradient descent (SGD) method is used to update all parameters including $W_{t}, \theta_t$, $\vec{\omega}$, etc.

However, the proposed feature embedding is effective in case of short term transactions but is extremely susceptible to longer series spanning many years.
In addition, the model training would become very slow and difficult to achieve convergence, resulting in the failure of feature embedding.

\section{Lifelong Learning Network}
\label{sec:lifelong-struc}

\subsection{Basic Model}

Unlike previous mainstream models of time series prediction that calculate the loss function and update parameters until last time step, our proposed lifetime learning approach will update parameters at each month.
Therefore, we divide house price set according to monthly-based subsets.
$y_t$ is the $t$-th set corresponding to embedded representation $\mathbb{H}_t$ of the house, and the overall time series is $\mathcal{Y} = \cup_{t=1}^{T} y_{t}$.
We record the number of traded houses with price labels each month as $N_t$.
The RMSE can be  formalized as:
\begin{align}\label{eq:LL-Loss}
    \mathcal{R}_{t}= \sqrt{\frac{1}{N_t}\sum_{j=1}^{N_t}(\hat{y}_{t}(j)-y_{t}(j))^{2}},
\end{align}
where $\hat{y}_{t}(j)$ and $y_{t}(j)$ refer to the predicted price and sold price of $j$-th house, respectively.

%  Long-life structure figure
\begin{figure}[t]
	\center
	\includegraphics[width=0.43\textwidth]{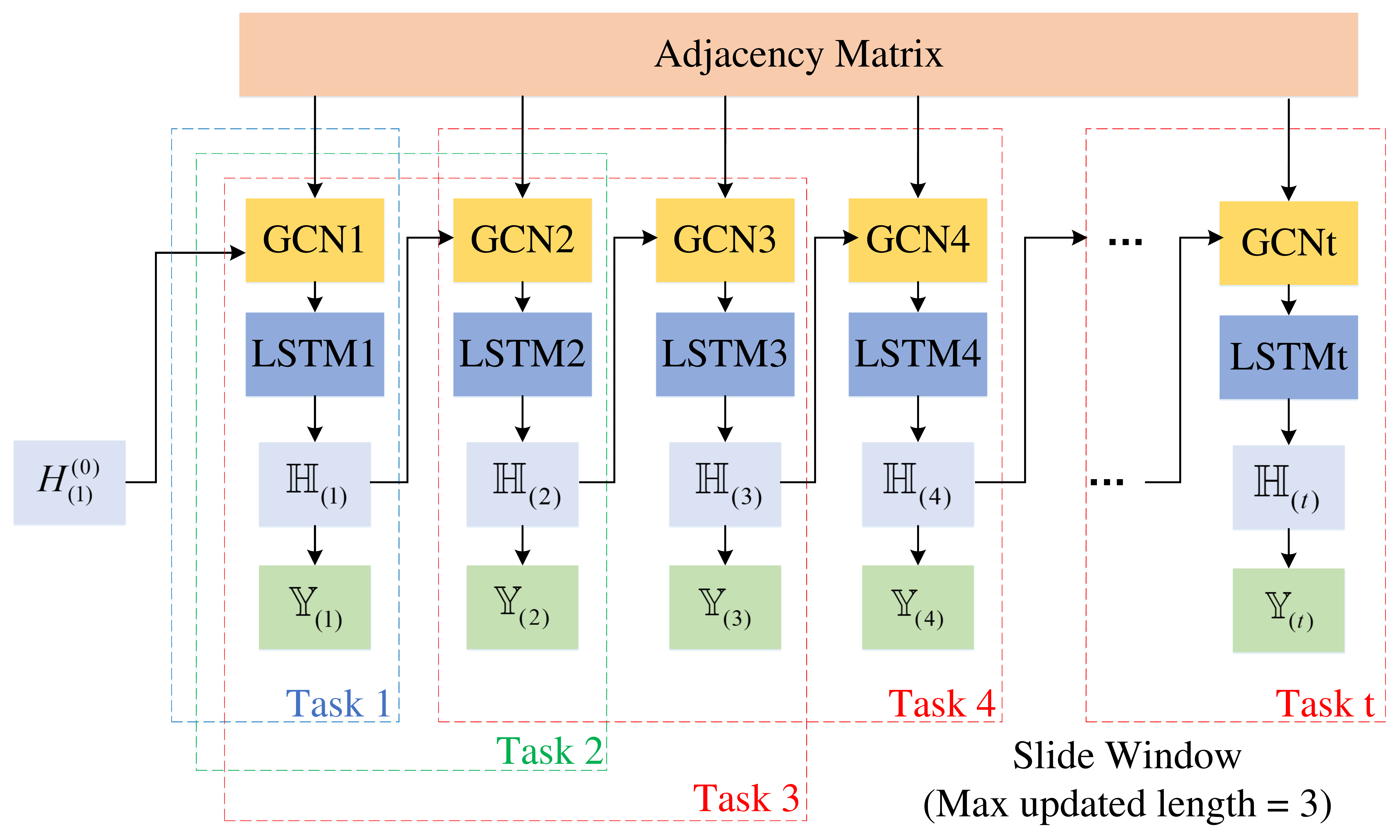}
	\vspace{-1em}
	\caption{An instance of the structure of \SystemName}\label{fig:lifelong}
	\vspace{-1em}
\end{figure}

In the lifelong learning network, we introduce a sliding recursive parameter updating strategy to navigate the depth of gradient back-propagation, thereby mitigating the gradient vanishing problem.
We take the evolving embedding $\mathbb{H}_{t}$ of the house learned every month as the initial house attribute of the next month's GCN-LSTM unit.
As shown in Fig.~\ref{fig:lifelong}, we use "$task$" to describe the training goal at various time steps in the lifelong model.
At time step $t$, task $t$ indicates the training procedure based on the $n$-months ahead of $t$ for the price label prediction at time step $t+1$, to confine the parameter updating for at most $n$ back-propagation depths.
For example, if $n$ is 3, within $Task$ $4$, the gradients generated by the objective function $\mathcal{R}_{4}$ will not only affect $W_4$ and $\theta_4$, but back-propagate and affect the parameters $W_3$, $\theta_3$ and $W_2$, $\theta_2$.

\mypara{Loss Function Calibration.} We differentiate the impacts of tasks on the effectiveness of feature learning.
Within each task, the loss function comprises the original loss function with the accumulation stemming from the back-propagation:
\begin{align}\label{eq:object_task}
\mathbb{L}_t = \frac{1}{n}\left(\mathcal{L}_{t} + \sum_{i=1}^{n-1}\lambda_{i}\Theta_{t-i}\right),
\end{align}
where $\lambda_{i}$ denotes the penalty coefficient to depict the impact of prior tasks to the task $t-1$ while $\Theta_{t-i}$ indicates the individual propagating loss.
To further accelerate the model training, we adopt \textit{parameter inheritance} in sequential GCN-LSTM units, which ensure the previously delivered house embedding can be initialized in the follow-up task.

%%Finally, $Task$ $t$ outputs the predicted value of house prices in the $t$-th month (step). Unlike traditional time series prediction models, an obvious advantage of \SystemName is that all houses in the entire city have very intuitive and different value predictions at different months (steps).

As only a small fraction of houses are traded every month and have price labels, when we calculate the monthly RMSE in Eq.~\ref{eq:LL-Loss}, only a part of the house is calculated, meaning $\hat{y}_{t}(j) \in \mathbb{Y}_t, j=1,2,\dots, N_t$.
Nevertheless, it is enough to train the proposed lifelong learning model.

\subsection{Reinforcement Learning based Optimization}
\label{sec:training}

Since the lifelong framework involves multi-tasks in the training procedure, it is indispensable to fine-tune the relevant coefficients as reasonable as possible.
To this end, we employ the reinforcement learning (RL) technique to facilitate the model optimization.

In Eq.~\ref{eq:object_task}, we collect the penalty coefficients as $\vec{\lambda} = [\lambda_1,\lambda_2\dots \lambda_{n-1}]$.
As instinctively transactions long time ago tend to have decayed impact on the up-to-date price prediction, we believe the $\lambda_{i}$ should become smaller when $task$ $t-i$ moves away from the current $task$ $t$.
We therefore use RL wherein the process of finding the minimum training loss is formalized as a Markov decision process (MDP) problem, i.e.,  $\mathcal{M}(\mathcal{S},\mathcal{A},P,r)$, where $\mathcal{S}$, $\mathcal{A}$, and $P$ are the state space, action space and state transition model, respectively.
$r$ represents the reward function:
\begin{itemize}[leftmargin=14pt]
    \item \textbf{State space:} The state $s\in\mathcal{S}$ is directly defined as the element in $\vec{\lambda}$: $s=\{\lambda_{0},\lambda_{1},\cdots,\lambda_{m}\}$, and $\lambda_{i}\in[0,1]$.

    \item \textbf{Action space:} The action $a\in\mathcal{A}$ is used to update the value of $\vec{\lambda}$. Since $\lambda_{i}\in[0,1]$, the action $a=(m,\epsilon)$ is to increase or decrease a small value $\epsilon$ for the $m$-th dimension in $\vec{\lambda}$, i.e., $\epsilon\in\{-0.01,0.01\}$.

    \item \textbf{Reward:} We determine whether $\vec{\lambda}$ is good enough by examining if the training loss calculated by the current $\vec{\lambda}$ can make the model achieve a smaller error on the test set, leading to a large delay in calculating reward $r$ according to the action $a$. To this end, we design it as a piece-wise function of discrete values, and directly use the discrete reward to update $\vec{\lambda}$ to simplify the RL process. In particular, we formalize the reward $r(s,a)$ as follows:
    \begin{align}\small
        r(s,a)=\left\{
        \begin{aligned}
            &+0.01,&if  & RMSE(\hat{y},y|s)>RMSE(\hat{y}',y|s')\\
            &-0.01,&if  & RMSE(\hat{y},y|s)<RMSE(\hat{y}',y|s')\\
        \end{aligned}
        \right.
    \end{align}
    where $RMSE (\hat{y}, y|s)$ represents the predicted price error of the model trained based on state $s$ on the test set, and $s'$ indicates the state after $s$ is updated according to action $a$.

    \item \textbf{State transition:} we determine the next state according to the reward $r(s,a)$ and the current action $a(m,\epsilon)$. If $r$ is positive, then the state will be transitted from state $s$ to $s'$ according to $a$; otherwise, state $s$ will remain the same.

    \item \textbf{Termination:} We make action $a$ randomly select the dimension $m$ to be updated, then $|\mathcal{A}|=2m$. If reward pertaining to each action is constantly negative within a range of certain steps, the loss coefficients are considered to be \textit{optimal}.
\end{itemize}

To facilitate the training of optimal loss coefficients, we, in practice, obtain the predicted price first through training on a small-scale dataset with fixed loss weight, before launching the RL. Once the model turns stable after repeated training iterations, we apply the loss weights directly to a larger dataset.

\section{Evaluation}\label{sec:eval}

\subsection{Experiment Setup}
\label{sec:dataset}
%\vspace{-0.2em}

\mypara{Platforms.} Evaluation is conducted on a multi-core server with a  64-core Intel Xeon CPU @2.40GHz with 512GB RAM and 8x NVIDIA
Tesla P100 GPUs. The server runs Ubuntu 20.04 LTS with Linux kernel 5.4.0. Our model is implemented using Python 3.5.2.

\mypara{Model parameters.} We set the house embedding dimension to $d=100$. The numbers of enumerated meta-paths and meta-graphs
participating in the calculation of house similarity are 30 and 80. We use month as the basic length of time steps to construct the time
series. We use Pytorch to implement the networks of the lifelong framework in \SystemName. We set 2 layers in GCN unit and we use ReLU as
the activation function. The size of hidden layers of LSTM is 128. We adopt Adam optimizer in the back propagation of the network, and the
learning rate is set to be 0.001 by default.

\begin{figure}[tb]	
	\centering
	\footnotesize
	\includegraphics[width=0.47\textwidth]{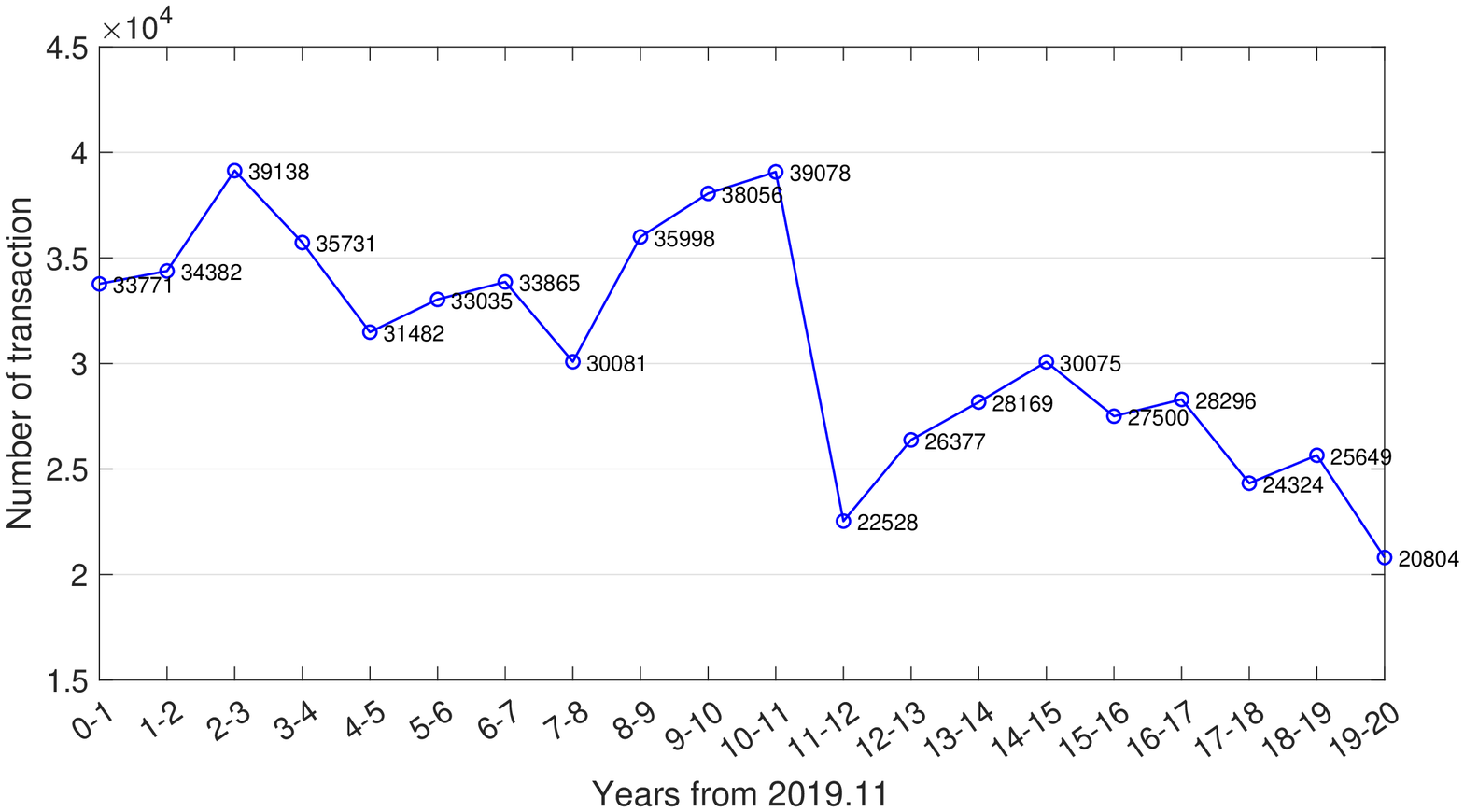}
	\vspace{-1em}
	\caption{The number of house transactions.}
	\label{fig:trans_num}
	\vspace{-1em}
\end{figure}

\mypara{Dataset.}
We sample houses in Toronto Region and construct multiple datasets.
We mainly use the following three datasets to evaluate our model:
\begin{itemize}[leftmargin=12pt]
    \item \textbf{TorC-H:} Sampled house transactions, spanning over 31 months, in a district in Toronto city.
    \item \textbf{TorC-A:} Sampled apartment transactions, spanning over 75 months, in a district in Toronto city. Apartment data has fewer features against houses.
    \item \textbf{Tor-H:} Sampled house transactions in the entire Toronto Region, which cover a wider area and  larger-scale data spanning over 63 months compared to the above two datasets. Fig.~\ref{fig:trans_num} depicts transaction numbers at different time intervals from November 2019.
\end{itemize}

%Dataset table
\begin{table}[t]
    \centering
    \caption{Statistics of the three datasets}\label{tab:dataset}
    \vspace{-1em}
    \setlength{\tabcolsep}{3mm}
    \small
    \begin{tabular}{l|rrr}
    \toprule
   \textbf{Dataset} & \textbf{TorC-H} & \textbf{TorC-A} & \textbf{Tor-H}\\
    \midrule
    \#House & 31,000 & 37,500  & 618,339  \\
    \#Nodes in \HINName & 155,834 & 127,955 & 3,093,180  \\
    Time span (months) & 31 & 75 & 246 \\
    \bottomrule
    \end{tabular}
\end{table}

Note that our sampling will make each house only have 1 transaction record.
Table \ref{tab:dataset} shows the statistic information of the three datasets.
Since TorC-H and TorC-A are relatively small-scale datasets, we mainly use them to learn penalty coefficients $\vec{\lambda}$, the weights of meta-paths and meta-graphs $\omega_{m}$, and evaluate the effectiveness of \SystemName, particularly comparing the predicted price and actual transaction price from 2019-06 to 2019-11.
We use the larger Tor-H to evaluate its capability of processing large-scale data.

\mypara{Evaluation metrics and baselines.} We mainly use the following two metrics in the experiments: (i) we use \textbf{RMSE} and
\textbf{MAE} to evaluate the error between the predicted price and the actual sold price of the house; (ii) we define \textbf{General Error
Rate (GER)} $R_{pre}$ as $R_{pre} = \frac{\hat{y}-y}{y}$, where $\hat{y}$ is the predicted price and $y$ is the actual transaction price. In
fact, the GER can better satisfy homeowners, buyers, agents and bank valuation agencies to understand the predicted
prices more intuitively.

%prediction distribution figure
\begin{figure*}[t]
	\centering	
	\subfigure[SVR]{
		\includegraphics[width=0.18\textwidth]{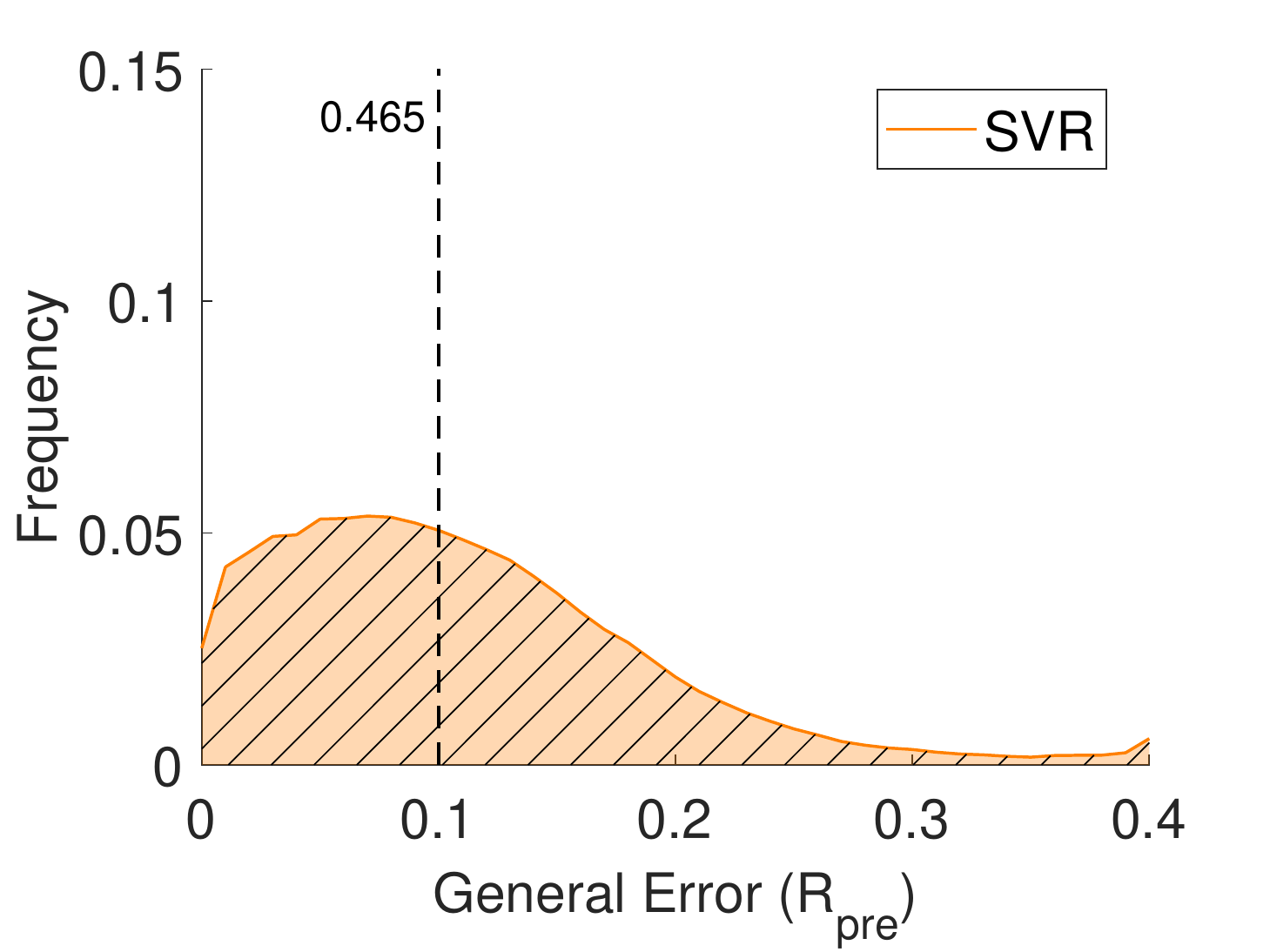}
		\setlength{\leftskip}{-20pt}
		\label{fig:predis_svr}
	}
	\subfigure[LSTM-D]{
		\includegraphics[width=0.18\textwidth]{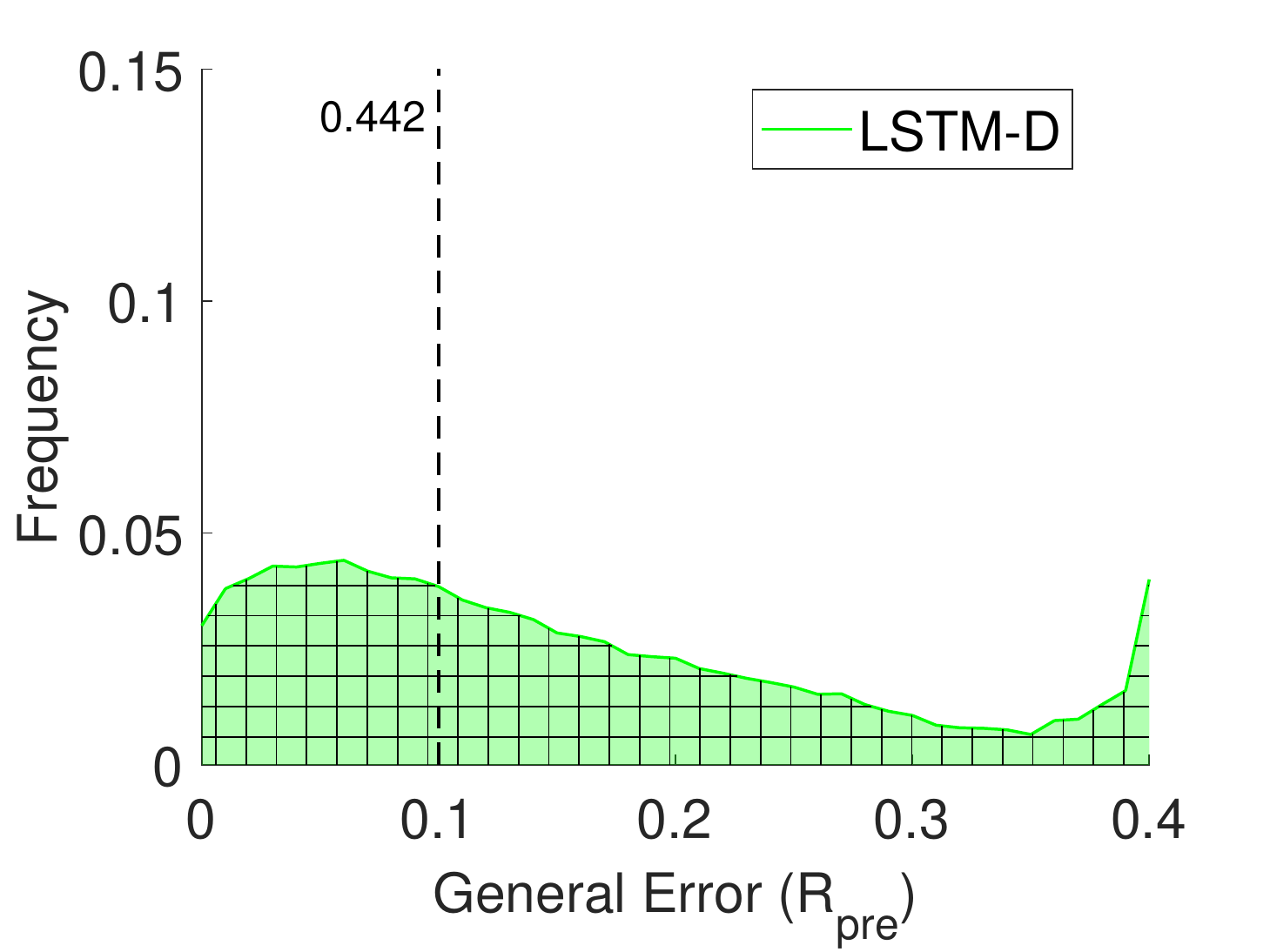}
		\setlength{\leftskip}{-20pt}
		\label{fig:predis_lstmd}
	}
	\subfigure[H-GCN]{
		\includegraphics[width=0.18\textwidth]{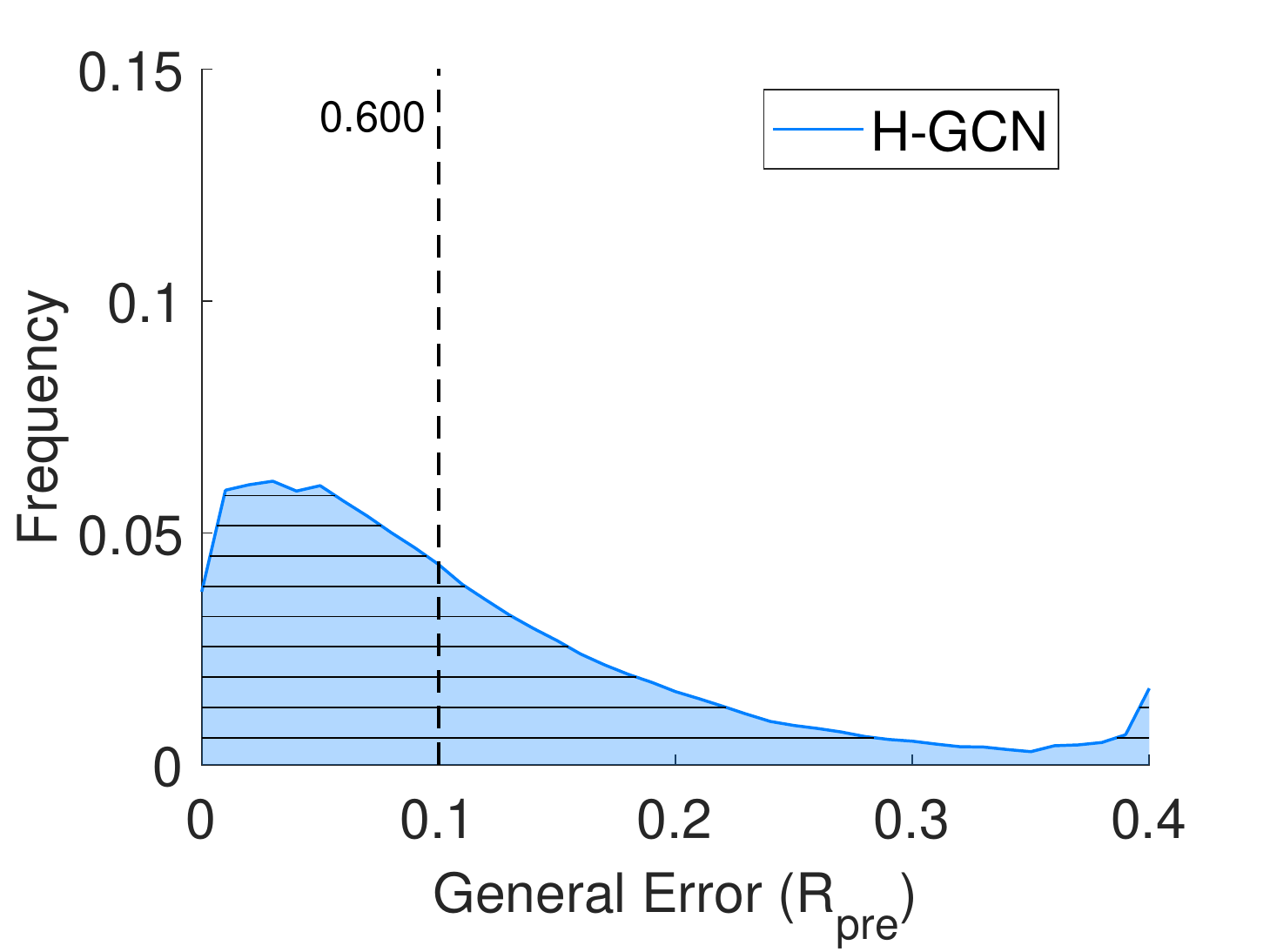}
		\setlength{\leftskip}{-20pt}
		\label{fig:predis_gcn}
	}
	\subfigure[HG-LSTM]{
		\includegraphics[width=0.18\textwidth]{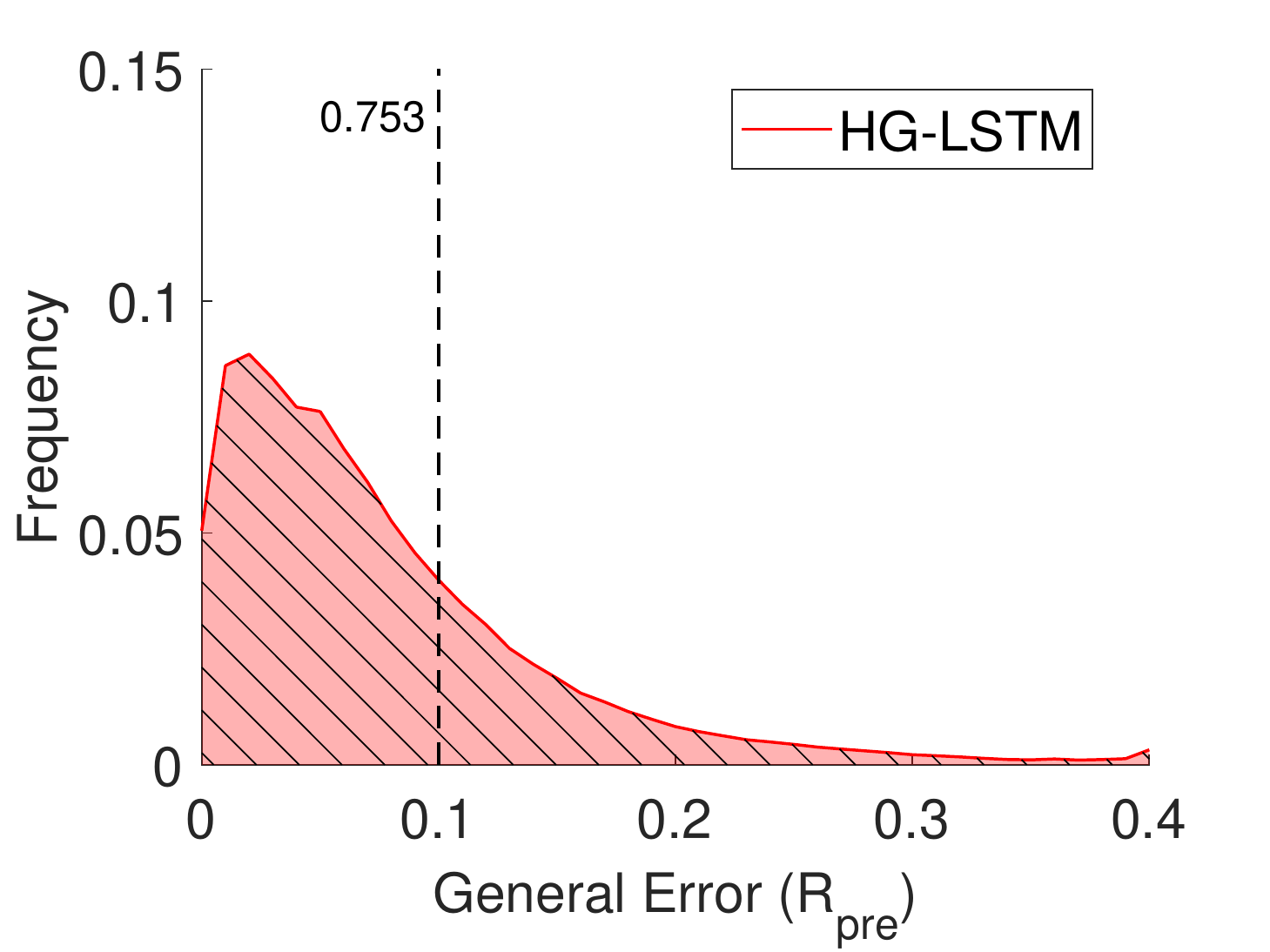}
		\setlength{\leftskip}{-20pt}
		\label{fig:predis_hglstm}
	}
	\subfigure[\SystemName]{
		\includegraphics[width=0.18\textwidth]{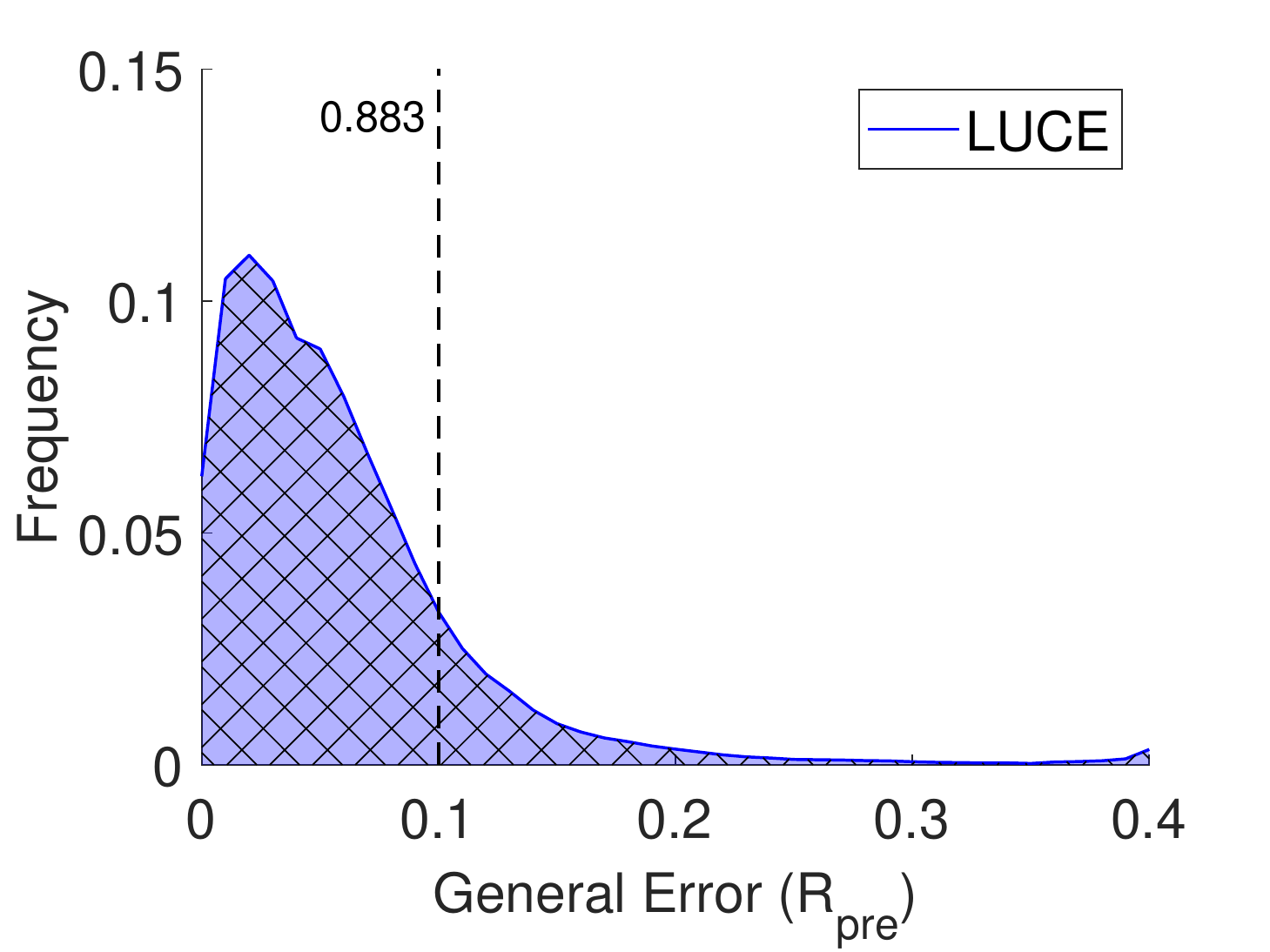}
		\setlength{\leftskip}{-20pt}
		\label{fig:predis_ll}
	}
	\vspace{-1em}
	\caption{The general error $R_{pre}$ distribution of house predicted price in TorC-H (2019-06 to 2019-11).}\label{fig:predis}
	\setlength{\leftskip}{-20pt}
	\vspace{-1em}
\end{figure*}

To evaluate the performance of the proposed \SystemName, we consider various of baseline methods as follows:
%\vspace{-0.3em}
\begin{itemize}[leftmargin=15pt]
    \item \textbf{Support vector regression (SVR)}: A classic machine learning algorithm that uses support vector machine (SVM) to fit the
        dataset for regression analysis. We set the input of the SVR as the initial feature of the houses, and utilize different SVR
        models to train the houses in different areas.
    \item \textbf{Decision tree (DT)}: A supervised machine learning method. Decision tree has a tree structure wherein each node
        represents the judgment of the attributes, and each branch represents the output of the judgment result. This is a
        non-time-series regression model.
    \item \textbf{Discrete time series based LSTM (LSTM-D)}: LSTM~\cite{hochreiter1997long} is a time-recurrent neural network model,
        which is widely used in temporal prediction. Since our housing transactions are not continuous in time, we organize the discrete
        housing initial features into time series and use them as input to the LSTM.
    \item \textbf{\HINName based temporal GCN (HT-GCN)}: An adaptive version from T-GCN~\cite{zhao2019t} where is a spatio-temporal
        prediction model combining GCN and GRU to extract spatial-temporal features. In our experiment, we use the adjacency matrix $A$
        generated by \HINName and the initial house attribute matrix $X$ as the inputs to T-GCN.
    \item \textbf{Heterogeneous GCN (H-GCN)}: H-GCN is consistent with the popular GCN unit~\cite{KipfSemi}, but only the spatial
        features of houses in \HINName are considered. This is a model of static semi-supervised learning on the constructed \HINName.
    \item \textbf{LSTM with H-GCN (HG-LSTM)}: HG-LSTM, consistent with the spatio-temporal feature learning unit mentioned in
        \S~\ref{sec:GCNLSTM}, is used for mining spatio-temporal features based on the constructed \HINName. It uses LSTM to explore
        temporal features of house prices without any lifelong framework. H-CGN provision input features for LSTM.
    \item \textbf{Independent expert valuation (Appraiser)}: We  give the estimated price based on the expertise of professional
        appraisers listed on the Toronto Real Estate Board~\cite{datasource}.
\end{itemize}

\mypara{Methodology.}
We mainly evaluate \SystemName in terms of the overall effectiveness and its breakdown stemming from system components and training optimizations.

We firstly evaluate the effectiveness of \SystemName on house price prediction by examining the general error rate and the average RMSE and MAE of different comparative methods against \SystemName (\S\ref{sec:effectiveness}). We further validate the effectiveness gained from our lifelong learning framework (\S\ref{sec:lifelong_eff}). Afterwards, we conduct in-depth investigations into the performance gain harvested from our design optimization and impact of parameters used in the lifelong learning component (\S\ref{sec:micro_eval}). In addition to these overall evaluation on effectiveness, we evaluate how regularization of training loss and parameter inheritance can boost the training effectiveness and accelerate the model training (\S\ref{sec:opt_eff}. We also outline our findings of digging and differentiating the importance in various meta-paths and meta-graphs and how they fundamentally underpin the effective feature embedding in GCNs (\S\ref{sec:meta_imp}).

As some baselines are difficult to deal with large-scale data, we split the large-scale dataset and train multiple identical models on the sub-datasets before merging their prediction results. For \SystemName, we make each GCN unit in the model process about 10,000 houses. When utilizing the above baseline methods to predict the transaction price of a month, all transaction prices of its previous months are used as training set to ensure the holistic test fairness.

%For an area, we calculate the number of houses at different general error levels as a percentage of all houses in the area. The general error is divided into four levels: $<5\%$, $<10\%$, $<15\%$ and $<20\%$.Since the transaction time of different houses scatters over different months, we test the traded houses in the last months in the experiments.

% prediction distribution table
\begin{table}[t]
    %\footnotesize
    \centering
    %\small
    \caption{General Error Rate ($R_{pre}$) Cumulative Probability in Different Methods (2019-11)}\label{tab:err-small}
    \vspace{-1em}
    \setlength{\tabcolsep}{2.5mm}
    \begin{tabular}{l|l|rrrr}
    \toprule
      &  \multirow{2}*{\textbf{Methods}}  &  \multicolumn{4}{c}{\textbf{GER Value}} \\
     & & $<5\%$ & $<10\%$ & $<15\%$ & $<20\%$ \\
    \midrule
    \multirow{8}*{\begin{sideways}TorC-H\end{sideways}}
    ~ & SVR & 23.72 & 46.47 & 66.75 & 78.50   \\
    ~ & DT & 29.71 & 55.60 & 72.84 & 83.34  \\
    ~ & LSTM-D & 22.95 & 44.18 & 61.32 & 75.89  \\
    \cmidrule{2-6}
    ~ & Appraiser & 46.84 & 68.61 & 83.78 & 91.56\\
    \cmidrule{2-6}
    ~ & H-GCN & 33.28 &  60.00 & 76.48 & 86.90  \\
    ~ & HT-GCN & 43.17 & 73.34 & 87.20 & 92.73  \\
    ~ & HG-LSTM & 47.93 & 75.28 & 88.38 & 94.13 \\
    \cmidrule{2-6}
    ~ & \SystemName & \textbf{60.44}  & \textbf{88.28}  & \textbf{95.44}  & \textbf{98.06} \\
    \midrule
    \multirow{8}*{\begin{sideways}TorC-A\end{sideways}}
    ~ & SVR & 23.47  & 44.29  & 61.89  &  77.18  \\
    ~ & DT & 25.83 & 48.47 & 66.10 & 77.67  \\
    ~ & LSTM-D & 22.95 & 43.27 & 61.64 & 75.58  \\
    \cmidrule{2-6}
    ~ & Appraiser & 37.80 & 54.41 & 70.60 & 83.66\\
    \cmidrule{2-6}
    ~ & H-GCN & 31.61  &  51.20 & 68.60  &  84.46  \\
    ~ & HT-GCN & 42.40 & 61.83 & 79.74 & 93.25  \\
    ~ & HG-LSTM & 45.62  & 66.42  & 84.22  &  95.47  \\
    \cmidrule{2-6}
    ~ & \SystemName & \textbf{52.84} &  \textbf{70.45}  & \textbf{90.81}  & \textbf{97.69} \\
    \midrule
    \multirow{8}*{\begin{sideways}Tor-H\end{sideways}}
    ~ & SVR & 19.60  & 38.36  & 54.91  &  67.05  \\
    ~ & DT & 24.42 & 44.05 & 62.48 & 74.36  \\
    ~ & LSTM-D & 27.85 & 48.58 & 64.83 & 76.87  \\
    \cmidrule{2-6}
    ~ & Appraiser & $\textbf{48.56}$ & 67.27 &87.13 & 96.60\\
    \cmidrule{2-6}
    ~ & H-GCN & 27.21 & 51.16 & 62.34 & 75.58  \\
    ~ & HT-GCN & 41.83 & 68.92 & 84.48 & 93.10  \\
    ~ & HG-LSTM & 36.04 & 65.75& 83.09 & 93.01  \\
    \cmidrule{2-6}
    ~ & \SystemName & 47.16 & $\textbf{68.07}$ & $\textbf{91.62}$ & $\textbf{97.38}$ \\
    \bottomrule
    \end{tabular}%\vspace{-0.5em}
\end{table}

\subsection{Prediction Effectiveness}
\label{sec:effectiveness}

\mypara{General error rate.} To illustrate the prediction effectiveness using transactions across multiple months, we firstly plot the
frequency distribution histogram of GER over a 6-months range from 2019-06 to 2019-11 in dataset TorC-H (containing a total of 6000 house
transaction records) under different representative methods\footnote{Since the techniques of some baseline methods used are similar, and
the different between their prediction results is small, we select SVR, LSTM-D, H-GCN, HG-LSTM, and \SystemName as the representatives.} in
Fig.~\ref{fig:predis}. We set the bin size to 0.01. Obviously, the shape with a left tendency indicates a lower error rate and better
prediction effectiveness.

It is observable that time series based approaches including HG-LSTM and \SystemName have much lower general error rate against other non-time-series approaches, e.g., H-GCN. Although HG-LSTM takes heterogeneous graph modeling and temporal features into account, its effectiveness is still inferior to \SystemName.
This is because we merely update parameters that have a more apparent and direct impact in limited depth, while other methods simply update all their parameters. \SystemName is also efficient in dealing with the prediction problem in case of long-term sparse transactions. To more precisely, we count the number of houses with GER lower than 10\% under various approaches. \SystemName has the highest number -- 47.13\% and 14.73\% more than H-GCN and HG-LSTM, respectively.

To have an in-depth understanding of the effectiveness, we dive into the experiment result of month 2019-11 and Table~\ref{tab:err-small} illustrates the corresponding cumulative proportion of houses with \textit{good prediction} (e.g., with less than 10\% GER) in all houses -- an important indicator of prediction precision used in real estate field.
The specific number indicates the percentage of houses that have GER less than a given threshold.
For instance, by using \SystemName on TorC-H dataset, 88.28\% houses can be \textit{perfectly} predicted the price with less than 10\% error rate. The number is far better than conventional SVR regressor (46.47\%) and appraiser-based estimation (68.61\%).

 %  Prediction Analysis Bar Chart
\begin{figure}[t]	
	\centering
	\subfigure[The RMSE of the prediction results on TorC-H]{
		\includegraphics[width=0.45\textwidth]{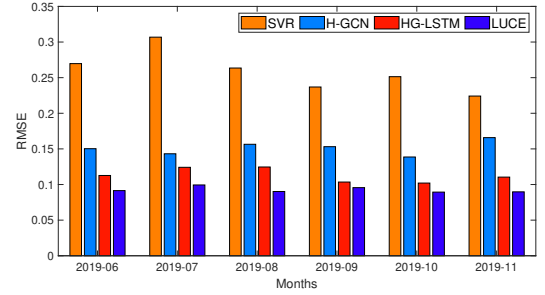}
		\label{fig:pre_ana_bar_rmse}
	}
	\subfigure[The MAE of the prediction results on TorC-H]{
		\includegraphics[width=0.45\textwidth]{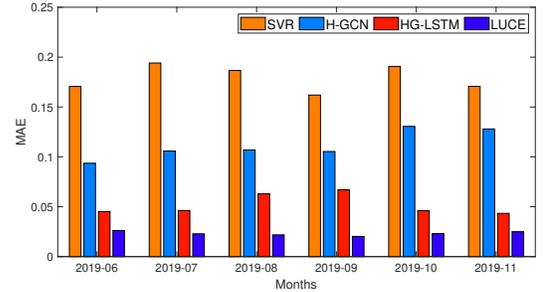}
		\label{fig:pre_ana_bar_mae}
	}
	\vspace{-1em}
	\caption{The RMSE and MAE of the prediction results on TorC-H in the last 6 continuous months (2019-06 to 2019-11).}\label{fig:pre_ana_bar}
	\vspace{-1em}
%	\vspace{-1em}
\end{figure}

\begin{table}[t!]
    \footnotesize
    \centering
    \caption{RMSE and MAE comparison (2019-11)}\label{tab:pre_rmse}
    \setlength{\tabcolsep}{5mm}
    \vspace{-1em}
    \begin{tabular}{l|l|rr}
    \toprule
    & \textbf{Methods} & \textbf{RMSE} & \textbf{MAE}\\
    \midrule
    \multirow{7}*{\begin{sideways}TorC-H\end{sideways}}
    ~ & SVR & 0.2852 & 0.1881 \\
    ~ & DT & 0.2794 & 0.1694 \\
    ~ & LSTM-D & 0.3051 & 0.2031 \\
    \cmidrule{2-4}
    ~ & Appraiser & 0.1158 & 0.0409\\
    \cmidrule{2-4}
    ~ & H-GCN & 0.1580 & 0.0958 \\
    ~ & HT-GCN & 0.2018 & 0.1349 \\
    ~ & HG-LSTM & 0.1333 & 0.0504 \\
    \cmidrule{2-4}
    ~ & \SystemName & \textbf{0.0950} & \textbf{0.0297} \\
    \midrule
    \multirow{7}*{\begin{sideways}TorC-A\end{sideways}}
    ~ & SVR & 0.3012 & 0.1964 \\
    ~ & DT & 0.3604 & 0.2230 \\
    ~ & LSTM-D & 0.3223 & 0.2255 \\
    \cmidrule{2-4}
    ~ & Appraiser & 0.1459 & 0.0941\\
    \cmidrule{2-4}
    ~ & H-GCN & 0.1984 & 0.1456 \\
    ~ & HT-GCN & 0.2575 & 0.1799 \\
    ~ & HG-LSTM & 0.1627 & 0.1068 \\
    \cmidrule{2-4}
    ~ & \SystemName & \textbf{0.1336} & \textbf{0.0742} \\
    \midrule
    \multirow{7}*{\begin{sideways}Tor-H\end{sideways}}
    ~ & SVR & 0.3185 & 0.2185 \\
    ~ & DT & 0.3001 & 0.1967 \\
    ~ & LSTM-D & 0.2893 & 0.1847 \\
    \cmidrule{2-4}
    ~ & Appraiser & 0.1677 & 0.1014 \\
    \cmidrule{2-4}
     ~ & H-GCN & 0.2105 & 0.1322 \\
     ~ & HT-GCN & 0.1735 & 0.1081 \\
    ~ & HG-LSTM & 0.1812 & 0.1104 \\
    \cmidrule{2-4}
    ~ & \SystemName & \textbf{0.1581} & \textbf{0.0982} \\
    \bottomrule
    \end{tabular}%\vspace{-0.5em}
\end{table}

\mypara{RMSE and MAE.}
Fig.~\ref{fig:pre_ana_bar} shows the RMSE and MAE in representative baseline methods, i.e., SVR, H-GCN and HG-LSTM against \SystemName on the TorC-H dataset.
As shown in Fig.~\ref{fig:pre_ana_bar}, \SystemName has much lower RMSE and MAE for house prediction in every month, with the lowest error fluctuation compared with other methods.
Specifically, compared with H-GCN and HG-LSTM, the average RMSE in \SystemName has decreased by up to 84.84\% and 38.11\%, respectively.
%Similarly, the MAE of monthly house prices predicted by \SystemName has decreased by up to 465.91\% and 230.72\%, respectively.
Similarly, the MAE of monthly house prices predicted by \SystemName has decreased to only 0.029.
%by up to 465.91\% and 230.72\%, respectively.
%\yry{change to decreased to only xx\%, by 466\% is not a good expression...}
Table ~\ref{tab:pre_rmse} demonstrates the detailed comparison if we only extract the data of 2019-11, also revealing the fact that \SystemName significantly outperforms other baseline methods.

This is because simple machine learning methods such as SVR and DT, only utilize the original features of houses, without employing heterogeneous information modeling to capture the intrinsic relationship and connections among houses. Due to the non-time-series techniques used in transactions modeling, their prediction turns out to be the worst in most cases.
Regarding static graph neural network approaches such as H-GCN that are based on heterogeneous modeling, graph neural network can obtain a fusion representation based on the relationships between houses and thus facilitate to overcome the freshness and sparsity problem to some degree. Nevertheless, H-GCN model neglects the temporal dependencies in the transaction data, resulting in a performance discrepancy compared with other improvements via heterogeneous graph modeling, such as HT-GCN and HG-LSTM.

Due to the ignorance of heterogeneous characteristics, LSTM-D delivers inferior results compared against other time-series methods that can capture and model heterogeneous data. The results are even worse than SVR and DT on both TorC-H and TorC-A datasets.
This is primarily because LSTM usually requires a large amount of high-quality data to underpin the feature learning.
In the case of sparse transactions, the prediction of LSTM-D barely outperforms the general regression models; when the number of reference houses is insufficient, LSTM-D is even inferior to general regression models in some scenarios.

By contrast, such approaches as HT-GCN, HG-LSTM and \SystemName elaborately consider the data sparsity by adopting heterogeneous graph embedding that can fully leverage house similarity and temporal dependency in the feature learning.
In fact, HT-GCN and HG-LSTM are able to deliver competitive results on both TorC-H and TorC-A datasets -- they can output equivalent or even better estimation compared to appraisers.
Nevertheless, when dealing with long-term prediction problem, the lack of up-to-date house prices and valuation has much more negative impact on the prediction results.
Furthermore, we observe that noise data stemming from the distant past house transactions has non-negligible impact on model performance and cause catastrophic forgetting problem.
In comparision, \SystemName employs the evolving multitask embedding to achieve constant parameter updates, resulting in a better effectiveness than HT-GCN and HG-LSTM. Our solution is even superior to the estimation by appraisers in most cases; the improvement can reach up to 20.21\% on prediction error rate.

\subsection{Lifelong Prediction Effectiveness}
\label{sec:lifelong_eff} 
\begin{figure}[tb]	
	\centering
	\includegraphics[width=0.45\textwidth]{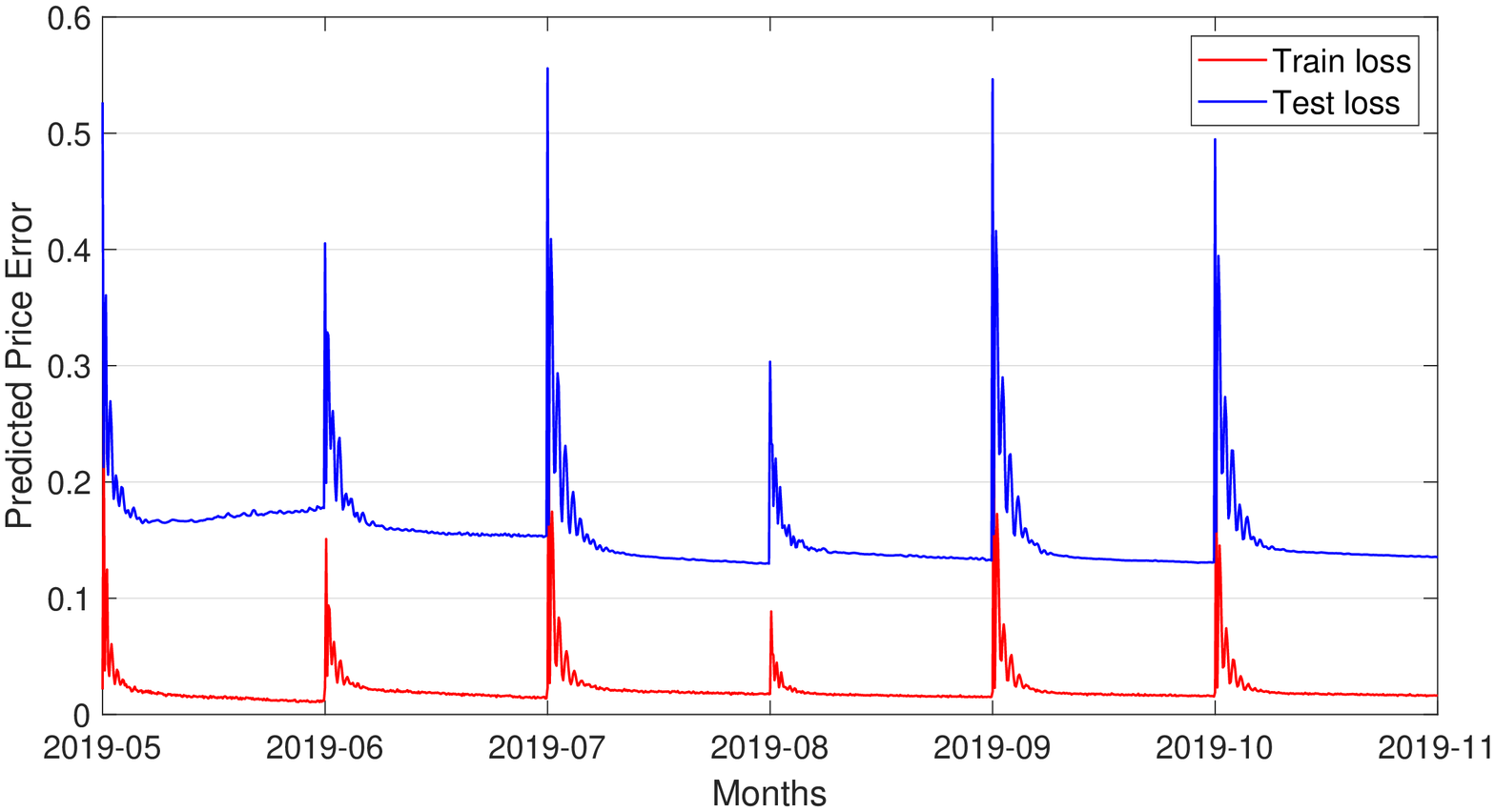}	
	\vspace{-1em}
	\caption{The train and test loss of \SystemName in simulating longlife continuous prediction}
	\label{fig:lifelong_error}
	\vspace{-1em}
\end{figure}

To further analyze the performance of handling large-scale and continuous prediction, we simulate \SystemName's
parameter update and prediction process in multi-month, on the occasion of new transaction data arrival from Tor-H. In this context, new
prices will be predicted by the model; once a house is traded, the transaction price will be added to the train set so that model parameter
will be updated. We record the loss of \SystemName on the train set and test set during the simulation, where the train set loss is
calculated by the optimized training loss in Eq. \ref{eq:object_task}, and the test set loss is calculated by RMSE. As shown in
Fig.~\ref{fig:lifelong_error}, the training procedure in \SystemName's is very stable across all stages -- the performance on the train set
and test set is generally consistent, indicating that continuous house price prediction can be effectively conducted in \SystemName on
large-scale datasets.

% Life-long error (6 month)

% Life-long price (6 month)
\begin{figure}[tb]	
	\centering
	\includegraphics[width=0.45\textwidth]{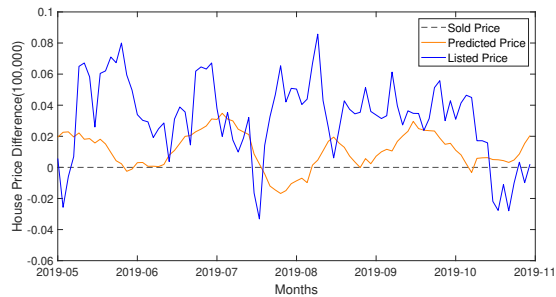}
	\vspace{-1em}
	\caption{Differences between predicted price and appraiser's price compared to actual sold price in Tor-H}
	\label{fig:lifelong_price}
	\vspace{-1em}
\end{figure}

To explicitly validate the effectiveness of lifelong prediction for the traded houses in the last six months of the dataset Tor-H, we plot in Fig.~\ref{fig:lifelong_price} the numerical discrepancies between the predicted price and the appraiser estimated price, together with the actual trading price as a baseline (in gray dotted line). Herein, we aggregate the average price of traded houses within an area.
Although the house prices vary over time in dataset Tor-H without any strong regularity, \SystemName is still able to deliver prices in closer proximity to the actual trading price compared with the appraiser's price.

To summarize, all the aforementioned experiments demonstrate that \SystemName has qualified learning capability of spatio-temporal features, and thus overcomes the data freshness and sparsity manifesting in house transaction records, against other baselines including simple regression methods and conventional spatio-temporal mining methods.

\subsection{Micro-benchmarking}
\label{sec:micro_eval}

In this section, we aim to demonstrate the individual contribution of different learning components to the holistic prediction effectiveness and performance.

% Ablation study figures
\begin{figure*}[t]
    \centering	
	\subfigure[RMSE comparison on TorC-H.]{
		\includegraphics[width=0.31\textwidth]{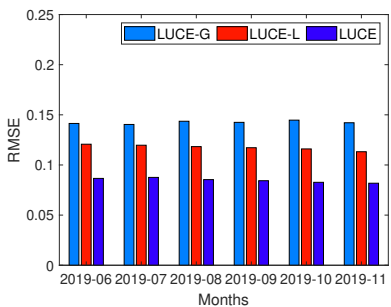}
		\setlength{\leftskip}{-20pt}
		\label{fig:ab_study_st1}
	}
	\subfigure[RMSE comparison on TorC-A.]{
		\includegraphics[width=0.31\textwidth]{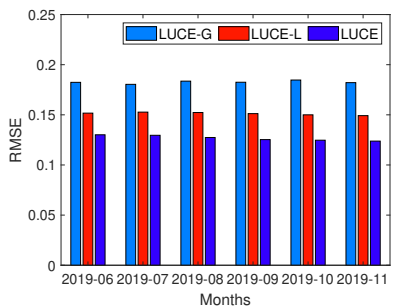}
		\setlength{\leftskip}{-20pt}
		\label{fig:ab_study_st2}
	}
	\subfigure[RMSE-maximum update lengths]{
		\includegraphics[width=0.31\textwidth]{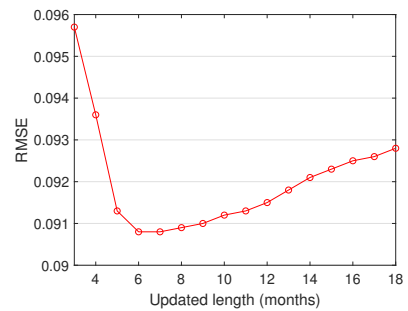}
		\setlength{\leftskip}{-20pt}
		\label{fig:ab_study_st3}
	}
	\vspace{-1em}
	\caption{The ablation study of \SystemName.}\label{fig:ab_study}
	\setlength{\leftskip}{-20pt}
	\vspace{-1em}
\end{figure*}

% Training Optimization figures
\begin{figure*}[t]
    \centering	
	\subfigure[Comparison of optimized loss and unoptimized loss on TorC-H.]{
		\includegraphics[width=0.31\textwidth]{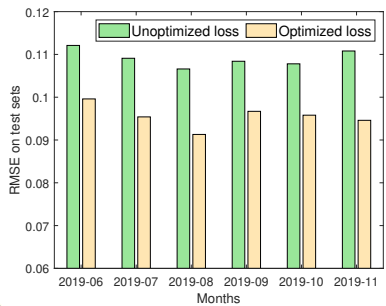}
		\setlength{\leftskip}{-20pt}
		\label{fig:mod_optim2}
	}
	\subfigure[Impact of parameter inheritance on TorC-H.]{
		\includegraphics[width=0.31\textwidth]{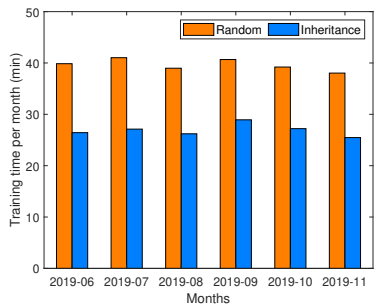}
		\setlength{\leftskip}{-20pt}
		\label{fig:mod_optim1}
	}
	\subfigure[Impact of the number of houses in one GCN.]{
		\includegraphics[width=0.31\textwidth]{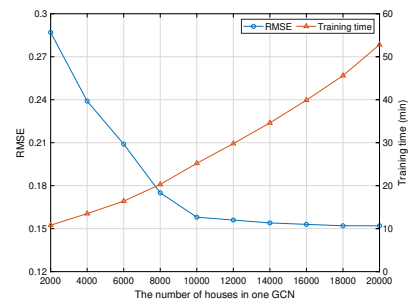}
		\setlength{\leftskip}{-20pt}
		\label{fig:mod_optim3}
	}
	\vspace{-1em}
	\caption{\SystemName's performance about training optimization}\label{fig:mod_optim}
	\setlength{\leftskip}{-20pt}
	\vspace{-1em}
\end{figure*}

\mypara{Ablation study: \SystemName-G, \SystemName-L vs. \SystemName.}  In light of the methodology of variable controlling, the main steps in this evaluation is to remain only one single component -- whilst removing others -- and examine how it affects the effectiveness. As depicted in Fig.~\ref{fig:lifelong} in \S\ref{sec:lifelong-struc}, \SystemName manages to continuously evolve the embedding by adaptively updating the parameters.
The updates mainly depend on GCN layers and LSTM layers while lifelong learning relies upon the combinations of such GCN-LSTM units and the limited-depth recursive parameter updating strategy.

Hence, we identify two comparable tailored subsystems -- \SystemName-G (\SystemName without LSTM layers) and \SystemName-L (\SystemName without GCN layers) -- and compare them with the complete \SystemName. We leverage the RMSE of the house prices prediction on the aforementioned 6-months transactions on  TorC-H (6,000 houses) and TorC-A (3,000 houses) dataset, respectively.
As shown in Fig.~\ref{fig:ab_study_st1} and Fig.~\ref{fig:ab_study_st2}, there is an RMSE increase in the
\SystemName-G and \SystemName-L compared against \SystemName, and the removal of LSTM layers has a greater impact on the performance than removal of GCN layers.
This indicates the proposed lifelong learning framework can  effectively and constantly tolerate the deficiency in up-to-date transaction data and alleviate the issue of time discontinuity.

\mypara{Impact of parameter in lifelong learning.}
We further investigate how the lifelong learning parameter  impact the overall performance of \SystemName. Specifically, the key variable is the number of spatio-temporal feature learning units, i.e, the maximum updated length $n$. We retain the same adoption of RL based training optimization and examine the RMSE
of predicting house prices in 2019-11 based on the dataset TorC-H. As shown in Fig.~\ref{fig:ab_study_st3}, the optional configuration of the  maximum update length is 6.
With the increment of length, the back-propagation tends to experience vanished gradient increasingly and more out-of-date transaction data will be involved in the learning. Overall, the model's prediction error could be acceptable when the maximum update length is between 5-12 months.
A smaller parameter will give rise to the surging RMSE because our model cannot completely explore the data of adjacent areas and months -- the model fails to learn the spatio-temporal features sufficiently.

This study implies it is extremely imperative to carry out the lifelong learning framework within \SystemName -- in each of the prediction tasks, we desire to train \SystemName with a moderate recursive length thereby effectively evolving the house embedding and minimizing the prediction error.

%In addition, when there are too many neural network unit layers in \SystemName, it will inevitably cause a huge training overhead.
%Therefore, although the impact caused by the large number of spatio-temporal feature learning units is not so great, from the perspective of model calculation cost, we do not advocate training a model with too many neural network unit layers.

%Figure \ref{fig:ab_study} shows the results of \SystemName's ablation study.
%First, we test the performance of the variation of \SystemName on datasets TorC-H and TorC-A.

%This indicates that \SystemName can fully mine the spatio-temporal features.
%Since \HINName establishes relationships for sparse reference houses, the adverse effects caused by removing the LSTM layer are greater than removing the GCN layer.
%And because \SystemName is a model for life-long learning in time series, its dependence on temporal features is stronger.
% Although the dependence on spatial features is not as strong as temporal features, the learning of spatial features can still greatly improve the accuracy of house price prediction.

%Then, we change the number of spatio-temporal feature learning units of \SystemName, that is, the maximum update length $n$, to verify the effectiveness of life-long learning framework.

%\vspace{-0.65em}
\subsection{Effectiveness of Training Optimization}
\label{sec:opt_eff}

In \S\ref{sec:GCNLSTM}, we regularize the house embedding in overlapping areas to optimize the training process of \SystemName. In this section, we present some optimization details during the training process and test the effectiveness of these optimizations during training process.

\mypara{Regularization of training loss.}
The optimization of training loss encompasses several portions including Eq.~\ref{eq:regularization}, Eq.~\ref{eq:object} and
Eq.~\ref{eq:object_task}. By contrast, the unoptimized training loss will be conducted without distance regulation, i.e., $\epsilon(P_t) =
0$\footnote{For this reason, we also use a separate multi-layer perception for each graph neural network to perform price prediction.}.

We therefore evaluate two cases where \SystemName is attached with and without such optimization based on 6-months TorC-H dataset, whilst
using RMSE as the main indicator. As depicted in Fig.~\ref{fig:mod_optim2}, \SystemName with regularization can significantly lower the
prediction error against \SystemName without regularization;  the RMSE value can be reduced by 11.15\% at most. This phenomenon is because
distance regularization can integrate the features learned by the same house in different graph neural networks, thereby better
coordinating and calibrating the feature embedding. In comparison, models without regularization have to learn the house's own features
without strong connections and fusions from external embedding results that can be reused.

% Figure of weights of metapath and metagraph
%\begin{figure}[t]	
%	\centering
%	\includegraphics[width=0.48\textwidth]{figures/Experiments/gcn_size.eps}\vspace{-1.5em}
%	\caption{Impact of house size in GCN.}
%	\label{fig:gcn_size}
%	\vspace{-1.5em}
%\end{figure}

\mypara{Parameters inheritance.} In order to shorten the time required for convergence during \SystemName training, we adopt the strategy of parameters inheritance and examine its efficiency.
%Both of these two techniques are parameter initialization methods proposed to enable the model to reach convergence faster.
%The pre-training is to use HG-LSTM to train the distant past data to obtain the initialization parameter values $W_{t}$, $\theta_{t}$ and $\vec{\omega}$ of $Task$ 1 of \SystemName.
This inheritance signifies the initial parameters $W_{t}$ and $\theta_{t}$ of a new time step $t$ can be possessed directly from the parameter $W_{t-1}$ and $\theta_{t-1}$ of its prior time step, without learning from the scratch. Intuitively, the inheritance takes advantage of similarities of evolving house embeddings in adjacent months, which is beneficial to the initialization of model parameters when new data arrives.

At the other extreme, parameters will be randomly initialized -- when the data of new time step $t$ arrives, the initial values of parameter $W_t$ and $\theta_t$ are given arbitrarily.
To evaluate the convergence time, we run
\SystemName models by using parameter inheritance (Inheritance) and random parameter initialization (Random), separately. we test their training time per month to achieve convergence on the dataset TorC-H in the last 6 month. Fig.~\ref{fig:mod_optim1} indicates that parameter inheritance can facilitate to reduce the training time; the training time required to achieve convergence can be reduced by up to 33.90\%. 
%We plot the training loss of \SystemName using the above optimization methods when training on the 6 months of TorC-H. As shown in Fig.~\ref{fig:mod_optim3}, the training loss tends to stabilize and continues to decrease with the continuous aggregation of new data.

\mypara{Scalability: impact of house number.}
We conduct experiments to examine the impact of varying the number of houses within a GCN on RMSE and training time required to reach convergence, by ranging the number from 2,000 to 20,000.  As depicted in Fig.~\ref{fig:mod_optim3}, when only a few houses available for learning in a GCN, it is inadequate for the GCN to effectively learn features of spatial information, due to the limited house overlap across different GCN units. Taking the dataset Tor-H as an example: the overlapping houses account for merely 4.73\% of all houses on average in a single GCN. 
By contrast, the increment of the total number of houses results in a soaring number of overlapping, thereby improving the effectiveness of distance regulation. However, the training overhead will grow drastically when dealing with a vast number of house transactions -- the training time to convergence increases significantly, susceptible to memory overflow in some worse-case scenarios. Hence,  we leverage a proper number of houses (i.e. 10,000 in the experiments) in building the graph and GCN, to ultimately balance the training time and precision requirement under memory constraints.

\subsection{Importance of Meta-paths and Meta-graphs}
\label{sec:meta_imp}

\begin{figure}[t]	
	\centering
	\includegraphics[width=0.5\textwidth]{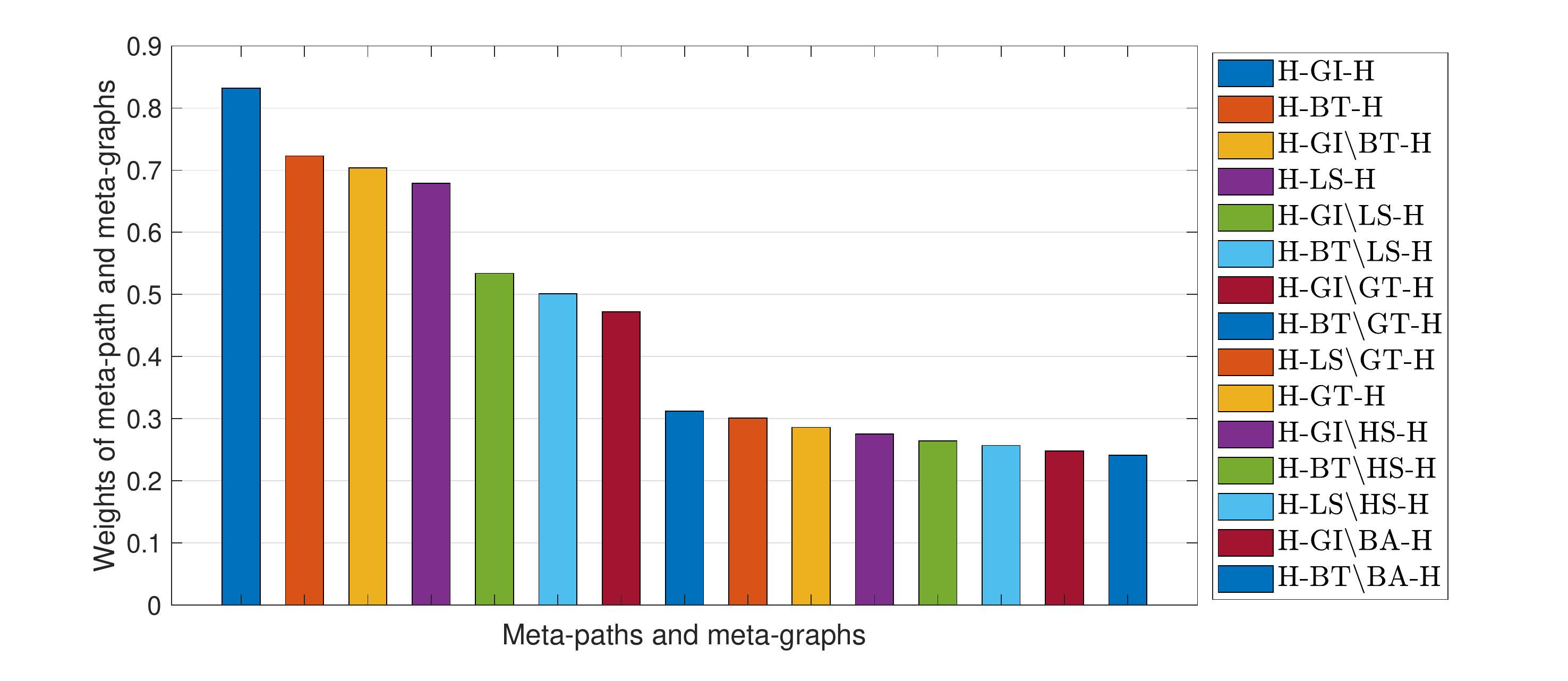}
	\vspace{-1em}
	\caption{Top 15 weights of meta-paths and meta-graphs.}
	\label{fig:meta_weight}
	\vspace{-1em}
\end{figure}

Fig.~\ref{fig:meta_weight} shows the learnable weights $\vec{\omega}$ of different meta-paths and meta-graphs after training on the dataset TorC-H. We display the top 15 weights of the delivered meta-paths and meta-graphs and observably there are  non-negligible differences in weights between different meta-paths and meta-graphs.
We can find House-Spatial Information-House (H-SI-H) is the meta-path with the largest weight, indicating that  the space information (SI) has the greatest impact on house prices among the various attributes in \HINName, followed by building type (BT), layout structure (LS), garage type (GT), and so forth.
This finding is coherent with our common understanding of property valuation.  Meanwhile, the disparity among meta-paths and meta-graphs make it reasonable to calculate the inherent similarity whilst recognizing the most crucial factors that have heavy impact on the real estate market.

% Figure of weights of metapath and metagraph

\section{Related Work}
\label{sec:relat}

\mypara{House price prediction.} The prediction of house prices attracts researchers' attention because it can be regarded as a regression problem when there is sufficient transaction and characteristic information of houses. There have been a lot of studies to predict housing
prices through simple machine learning techniques (such as decision tree and hedonic model)~\cite{park2015using,bin2004prediction,
chica2007prediction, osland2010application, bourassa2007spatial} or deep learning neural networks with relatively simple
structures~\cite{limsombunchai2004house, selim2009determinants}. These techniques can generally take into account the spatial
characteristics of the houses, encode the characteristics and send them to the model for training.
%Based on the spatial information of the house, some researches\cite{montero2018housing,liu2016house} have also considered other characteristics of the house, such as surrounding facilities, house configuration, etc.
%These researches are aware of the heterogeneity of housing characteristics.
%Few studies have been able to model the houses characteristics as heterogeneous information networks and explore the similarity between houses.
In recent years, some researches~\cite{Bollerslev2016Daily, liu2013spatial} have considered the impact of temporal features on house prices, while fully considering other houses characteristics, and using time series models to predict housing prices.
% Therefore, for the problem of housing price prediction, a comprehensive method should be hierarchical heterogeneous spatio-temporal features modeling\cite{Tan2017Time}.

\mypara{Heterogeneous graph learning.} Here we mainly refer to the representation learning of heterogeneous graphs. Technically, it mainly
includes two types of unsupervised heterogeneous information network
embeddings~\cite{Fu2017HIN,Dong2017Metapath,Zhang2018Metagraph,He2019Hete,Ji2018Attention,Cen2019Repre,dos2016multilabel} and
semi-supervised heterogeneous graph neural networks~\cite{Wang2019Heter,Zhang2019Heter,Peng2019Fine}. In terms of unsupervised heterogeneous
information network embeddings, most of the approaches are based on meta-path~\cite{Fu2017HIN,Dong2017Metapath} or
meta-graph~\cite{Zhang2018Metagraph} guided random walk on heterogeneous network to learn the embedding of nodes with negative sampling
technologies.
%On this basis, some researches have introduced attention mechanisms\cite{Ji2018Attention} or upgraded the strategy of random walk\cite{He2019Hete}. These techniques are highly dependent on meta-path and meta-graph.
In terms of semi-supervised heterogeneous graph neural networks, most existing researches are based on homogeneous graph neural networks, fusing different types node information~\cite{Wang2019Heter,Zhang2019Heter} or converting heterogeneous graphs into parameterized homogeneous graphs~\cite{Peng2019Fine}, and then learning node embedding through graph neural networks.

\mypara{Spatio-temporal data mining.} Recent studies on spatio-temporal data prediction have combined models that extract spatial and
temporal features. For example, ConvLSTM~\cite{Shi2015Conv} is a combination of CNN and LSTM. In terms of spatial features, CNN is usually
used for images, maps or data that can be modeled as grids~\cite{DABIRI2018360,Chen2018Exploiting,Tran_2015_ICCV}, and graph neural network
is usually used for data that can be modeled as graphs and networks~\cite{Chai2018Bike,geng2019spatiotemporal,lin2018exploiting}. In terms
of temporal features, most researches utilize RNN to learn temporal features, including LSTM~\cite{akbari2018short,dixon2019deep},
GRU~\cite{zhao2019t}, Seq2Seq~\cite{liao2018dest}, and so on.

\mypara{Lifelong learning.} Lifelong learning is a relatively new research domain proposed in recent years, aiming to propose a method
that can accumulate past knowledge and apply it to future learning~\cite{chen2016lifelong}. In recent years, \cite{liu2018rotate} retained
the useful parameters for new tasks by changing the gradient update strategy, while ignoring those useless
parameters~\cite{rusu2016progressive} expanded the models and combines the trained models with the new model to train new tasks.
\cite{aljundi2017expert} designed the gate to determine which past task the new task is more like to initialize the model of the new task.
There are still a lot of works to study in the field of lifelong learning.

\section{Conclusion}\label{sec:conclu}
We have presented \SystemName, a novel learning framework for automated property valuation. \SystemName is designed to address the spatial
and temporal sparsity of house transaction data. To extract useful information, \SystemName organizes the house-related data in a
heterogeneous information network (HIN). It then employs the GCN and LSTM to extract the spatial and temporal information from the HIN.
\SystemName uses GCN and LSTM to develop a lifelong learning framework for house valuation for the first time. \SystemName makes use of the limited recent house transactions data to update the valuation for all house entities in the HIN to provide a
complete and update-to-date dataset to improve the accuracy of the downstream price prediction task. We evaluate \SystemName by applying it
to large-scale, real-world house transaction data of Toronto between 2000 and 2019. Experimental results show that \SystemName consistently
outperforms prior automated house valuation methods. It reaches and often exceeds the accuracy of valuation given by independent experts
when using the actual sold price as the ground truth.

%\section*{Acknowledgments}
%This work is supported in part by the National Key R\&D Program of China (2018YFC0830804), NSFC (61872022 and 61872294), the UK EPSRC (EP/T01461X/1), a UK Royal Society International Collaboration Grant, NSF (III-1526499, III-1763325, III-1909323), CNS-1930941, and NSF of Guangdong Province (2017A030313339).

\bibliography{reference}

\begin{thebibliography}{10}
\providecommand{\url}[1]{#1}
\csname url@samestyle\endcsname
\providecommand{\newblock}{\relax}
\providecommand{\bibinfo}[2]{#2}
\providecommand{\BIBentrySTDinterwordspacing}{\spaceskip=0pt\relax}
\providecommand{\BIBentryALTinterwordstretchfactor}{4}
\providecommand{\BIBentryALTinterwordspacing}{\spaceskip=\fontdimen2\font plus
\BIBentryALTinterwordstretchfactor\fontdimen3\font minus
  \fontdimen4\font\relax}
\providecommand{\BIBforeignlanguage}[2]{{%
\expandafter\ifx\csname l@#1\endcsname\relax
\typeout{** WARNING: IEEEtran.bst: No hyphenation pattern has been}%
\typeout{** loaded for the language `#1'. Using the pattern for}%
\typeout{** the default language instead.}%
\else
\language=\csname l@#1\endcsname
\fi
#2}}
\providecommand{\BIBdecl}{\relax}
\BIBdecl

\bibitem{bourassa2007spatial}
S.~C. Bourassa, E.~Cantoni, and M.~Hoesli, ``Spatial dependence, housing
  submarkets, and house price prediction,'' \emph{The Journal of Real Estate
  Finance and Economics}, vol.~35, no.~2, pp. 143--160, 2007.

\bibitem{basu1998analysis}
S.~Basu and T.~G. Thibodeau, ``Analysis of spatial autocorrelation in house
  prices,'' \emph{The Journal of Real Estate Finance and Economics}, vol.~17,
  no.~1, pp. 61--85, 1998.

\bibitem{park2015using}
B.~Park and J.~K. Bae, ``Using machine learning algorithms for housing price
  prediction: The case of fairfax county, virginia housing data,'' \emph{Expert
  Systems with Applications}, no.~6, pp. 2928--2934, 2015.

\bibitem{stevenson2004new}
S.~Stevenson, ``New empirical evidence on heteroscedasticity in hedonic housing
  models,'' \emph{Journal of Housing Economics}, vol.~13, no.~2, pp. 136--153,
  2004.

\bibitem{datasource}
\BIBentryALTinterwordspacing
Toronto real estate board. [Online]. Available: \url{http://trreb.ca/}
\BIBentrySTDinterwordspacing

\bibitem{bengio1994learning}
Y.~Bengio, P.~Simard, and P.~Frasconi, ``Learning long-term dependencies with
  gradient descent is difficult,'' \emph{IEEE transactions on neural networks},
  vol.~5, no.~2, pp. 157--166, 1994.

\bibitem{mccloskey1989catastrophic}
M.~McCloskey and N.~J. Cohen, ``Catastrophic interference in connectionist
  networks: The sequential learning problem,'' in \emph{Psychology of learning
  and motivation}.\hskip 1em plus 0.5em minus 0.4em\relax Elsevier, 1989, pp.
  109--165.

\bibitem{sun2012mining}
Y.~Sun, J.~Han, X.~Yan, and P.~S. Yu, ``Mining knowledge from interconnected
  data: a heterogeneous information network analysis approach,''
  \emph{Proceedings of the VLDB Endowment}, vol.~5, no.~12, pp. 2022--2023,
  2012.

\bibitem{shi2016survey}
C.~Shi, Y.~Li, J.~Zhang, Y.~Sun, and P.~S. Yu, ``A survey of heterogeneous
  information network analysis,'' \emph{IEEE Transactions on Knowledge and Data
  Engineering}, vol.~29, no.~1, pp. 17--37, 2016.

\bibitem{VeliGraph}
P.~Veli{\v{c}}kovi{\'{c}}, G.~Cucurull, A.~Casanova, A.~Romero, P.~Li{\`{o}},
  and Y.~Bengio, ``Graph attention networks,'' in \emph{Proceedings of the
  ICLR}, 2018.

\bibitem{wu2020comprehensive}
Z.~Wu, S.~Pan, F.~Chen, G.~Long, C.~Zhang, and P.~S. Yu, ``A comprehensive
  survey on graph neural networks,'' \emph{IEEE Transactions on Neural Networks
  and Learning Systems}, 2020.

\bibitem{Peng2019Fine}
H.~Peng, J.~Li, Q.~Gong, Y.~Song, Y.~Ning, K.~Lai, and P.~S. Yu, ``Fine-grained
  event categorization with heterogeneous graph convolutional networks,'' in
  \emph{Proceedings of the IJCAI}.\hskip 1em plus 0.5em minus 0.4em\relax AAAI
  Press, 2019, pp. 3238--3245.

\bibitem{KipfSemi}
T.~N. Kipf and M.~Welling, ``Semi-supervised classification with graph
  convolutional networks,'' in \emph{Proceedings of the ICLR}, 2017.

\bibitem{hochreiter1997long}
S.~Hochreiter and J.~Schmidhuber, ``Long short-term memory,'' \emph{Neural
  computation}, vol.~9, no.~8, pp. 1735--1780, 1997.

\bibitem{zhao2019t}
L.~Zhao, Y.~Song, C.~Zhang, Y.~Liu, P.~Wang, T.~Lin, M.~Deng, and H.~Li,
  ``T-gcn: A temporal graph convolutional network for traffic prediction,''
  \emph{IEEE Transactions on Intelligent Transportation Systems}, 2019.

\bibitem{bin2004prediction}
O.~Bin, ``A prediction comparison of housing sales prices by parametric versus
  semi-parametric regressions,'' \emph{Journal of Housing Economics}, vol.~13,
  no.~1, pp. 68--84, 2004.

\bibitem{chica2007prediction}
J.~Chica-Olmo, ``Prediction of housing location price by a multivariate spatial
  method: Cokriging,'' \emph{Journal of Real Estate Research}, vol.~29, no.~1,
  pp. 91--114, 2007.

\bibitem{osland2010application}
L.~Osland, ``An application of spatial econometrics in relation to hedonic
  house price modeling,'' \emph{Journal of Real Estate Research}, vol.~32,
  no.~3, pp. 289--320, 2010.

\bibitem{limsombunchai2004house}
V.~Limsombunchai, ``House price prediction: hedonic price model vs. artificial
  neural network,'' in \emph{New Zealand agricultural and resource economics
  society conference}, 2004, pp. 25--26.

\bibitem{selim2009determinants}
H.~Selim, ``Determinants of house prices in turkey: Hedonic regression versus
  artificial neural network,'' \emph{Expert systems with Applications},
  vol.~36, no.~2, pp. 2843--2852, 2009.

\bibitem{Bollerslev2016Daily}
T.~Bollerslev, A.~J. Patton, and W.~Wang, ``Daily house price indices:
  Construction, modeling, and longer-run predictions,'' \emph{Journal of
  Applied Econometrics}, vol.~31, no.~6, pp. 1005--1025, 2016.

\bibitem{liu2013spatial}
X.~Liu, ``Spatial and temporal dependence in house price prediction,''
  \emph{The Journal of Real Estate Finance and Economics}, vol.~47, no.~2, pp.
  341--369, 2013.

\bibitem{Fu2017HIN}
T.~Y. Fu, W.~C. Lee, and Z.~Lei, ``Hin2vec: Explore meta-paths in heterogeneous
  information networks for representation learning,'' in \emph{Proceedings of
  the ACM CIKM}, ser. CIKM '17, New York, NY, USA, 2017, p. 1797–1806.

\bibitem{Dong2017Metapath}
Y.~Dong, N.~V. Chawla, and A.~Swami, ``Metapath2vec: Scalable representation
  learning for heterogeneous networks,'' in \emph{Proceedings of the ACM
  SIGKDD}, 2017, p. 135–144.

\bibitem{Zhang2018Metagraph}
D.~Zhang, J.~Yin, X.~Zhu, and C.~Zhang, ``Metagraph2vec: Complex semantic path
  augmented heterogeneous network embedding,'' in \emph{Proceedings of the
  PAKDD}.\hskip 1em plus 0.5em minus 0.4em\relax Springer, 2018, pp. 196--208.

\bibitem{He2019Hete}
Y.~He, Y.~Song, J.~Li, C.~Ji, J.~Peng, and H.~Peng, ``Hetespaceywalk: A
  heterogeneous spacey random walk for heterogeneous information network
  embedding,'' in \emph{Proceedings of the ACM CIKM}, 2019, p. 639–648.

\bibitem{Ji2018Attention}
H.~Ji, C.~Shi, and B.~Wang, ``Attention based meta path fusion for
  heterogeneous information network embedding,'' in \emph{Proceedings of the
  PRICAI}, X.~Geng and B.-H. Kang, Eds., 2018.

\bibitem{Cen2019Repre}
Y.~Cen, X.~Zou, J.~Zhang, H.~Yang, J.~Zhou, and J.~Tang, ``Representation
  learning for attributed multiplex heterogeneous network,'' in
  \emph{Proceedings of the ACM SIGKDD}.\hskip 1em plus 0.5em minus 0.4em\relax
  New York, NY, USA: Association for Computing Machinery, 2019, p. 1358–1368.

\bibitem{dos2016multilabel}
L.~Dos~Santos, B.~Piwowarski, and P.~Gallinari, ``Multilabel classification on
  heterogeneous graphs with gaussian embeddings,'' in \emph{Proceedings of the
  ECMLKDD}, 2016, pp. 606--622.

\bibitem{Wang2019Heter}
X.~Wang, H.~Ji, C.~Shi, B.~Wang, Y.~Ye, P.~Cui, and P.~S. Yu, ``Heterogeneous
  graph attention network,'' in \emph{Proceedings of the WWW}, 2019, p.
  2022–2032.

\bibitem{Zhang2019Heter}
C.~Zhang, D.~Song, C.~Huang, A.~Swami, and N.~V. Chawla, ``Heterogeneous graph
  neural network,'' in \emph{Proceedings of the ACM SIGKDD}.\hskip 1em plus
  0.5em minus 0.4em\relax New York, NY, USA: Association for Computing
  Machinery, 2019, p. 793–803.

\bibitem{Shi2015Conv}
X.~SHI, Z.~Chen, H.~Wang, D.~Y. Yeung, W.-k. Wong, and W.~C. WOO,
  ``Convolutional lstm network: A machine learning approach for precipitation
  nowcasting,'' in \emph{Proceedings of the NIPS}, C.~Cortes, N.~D. Lawrence,
  D.~D. Lee, M.~Sugiyama, and R.~Garnett, Eds., 2015, pp. 802--810.

\bibitem{DABIRI2018360}
S.~Dabiri and K.~Heaslip, ``Inferring transportation modes from gps
  trajectories using a convolutional neural network,'' \emph{Transportation
  research part C: emerging technologies}, vol.~86, pp. 360--371, 2018.

\bibitem{Chen2018Exploiting}
C.~{Chen}, K.~{Li}, S.~G. {Teo}, G.~{Chen}, X.~{Zou}, X.~{Yang}, R.~C. {Vijay},
  J.~{Feng}, and Z.~{Zeng}, ``Exploiting spatio-temporal correlations with
  multiple 3d convolutional neural networks for citywide vehicle flow
  prediction,'' in \emph{Proceedings of the IEEE ICDM}, 2018, pp. 893--898.

\bibitem{Tran_2015_ICCV}
D.~Tran, L.~Bourdev, R.~Fergus, L.~Torresani, and M.~Paluri, ``Learning
  spatiotemporal features with 3d convolutional networks,'' in
  \emph{Proceedings of the IEEE ICCV}, December 2015.

\bibitem{Chai2018Bike}
D.~Chai, L.~Wang, and Q.~Yang, ``Bike flow prediction with multi-graph
  convolutional networks,'' in \emph{Proceedings of the ACM SIGSPATIAL}, New
  York, NY, USA, 2018, p. 397–400.

\bibitem{geng2019spatiotemporal}
X.~Geng, Y.~Li, L.~Wang, L.~Zhang, Q.~Yang, J.~Ye, and Y.~Liu, ``Spatiotemporal
  multi-graph convolution network for ride-hailing demand forecasting,'' in
  \emph{Proceedings of the AAAI}, vol.~33, 2019, pp. 3656--3663.

\bibitem{lin2018exploiting}
Y.~Lin, N.~Mago, Y.~Gao, Y.~Li, Y.-Y. Chiang, C.~Shahabi, and J.~L. Ambite,
  ``Exploiting spatiotemporal patterns for accurate air quality forecasting
  using deep learning,'' in \emph{Proceedings of the ACM SIGSPATIAL}, 2018, pp.
  359--368.

\bibitem{akbari2018short}
A.~Akbari~Asanjan, T.~Yang, K.~Hsu, S.~Sorooshian, J.~Lin, and Q.~Peng,
  ``Short-term precipitation forecast based on the persiann system and lstm
  recurrent neural networks,'' \emph{Journal of Geophysical Research:
  Atmospheres}, vol. 123, no.~22, pp. 12--543, 2018.

\bibitem{dixon2019deep}
M.~F. Dixon, N.~G. Polson, and V.~O. Sokolov, ``Deep learning for
  spatio-temporal modeling: Dynamic traffic flows and high frequency trading,''
  \emph{Applied Stochastic Models in Business and Industry}, vol.~35, no.~3,
  pp. 788--807, 2019.

\bibitem{liao2018dest}
B.~Liao, J.~Zhang, M.~Cai, S.~Tang, Y.~Gao, C.~Wu, S.~Yang, W.~Zhu, Y.~Guo, and
  F.~Wu, ``Dest-resnet: A deep spatiotemporal residual network for hotspot
  traffic speed prediction,'' in \emph{Proceedings of the ACM MM}, 2018, pp.
  1883--1891.

\bibitem{chen2016lifelong}
Z.~Chen, E.~R. Hruschka~Jr, and B.~Liu, ``Lifelong machine learning and
  computer reading the web,'' in \emph{Proceedings of the ACM SIGKDD}, 2016,
  pp. 2117--2118.

\bibitem{liu2018rotate}
X.~Liu, M.~Masana, L.~Herranz, J.~Van~de Weijer, A.~M. Lopez, and A.~D.
  Bagdanov, ``Rotate your networks: Better weight consolidation and less
  catastrophic forgetting,'' in \emph{Proceedings of the IEEE ICPR}, 2018, pp.
  2262--2268.

\bibitem{rusu2016progressive}
A.~A. Rusu, N.~C. Rabinowitz, G.~Desjardins, H.~Soyer, J.~Kirkpatrick,
  K.~Kavukcuoglu, R.~Pascanu, and R.~Hadsell, ``Progressive neural networks,''
  \emph{arXiv preprint arXiv:1606.04671}, 2016.

\bibitem{aljundi2017expert}
R.~Aljundi, P.~Chakravarty, and T.~Tuytelaars, ``Expert gate: Lifelong learning
  with a network of experts,'' in \emph{Proceedings of the IEEE CVPR}, 2017,
  pp. 3366--3375.

\end{thebibliography}

%\vskip -1.6\baselineskip
\begin{IEEEbiography}
[{\includegraphics[width=1in,height=1.25in,clip,keepaspectratio]{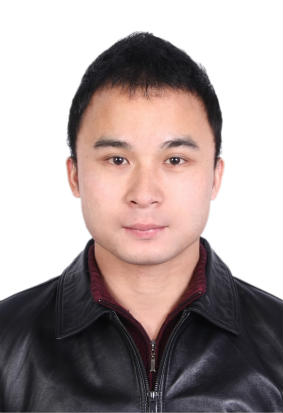}}]
{Hao Peng} is currently an Assistant Professor at the School of Cyber Science and Technology, and Beijing Advanced Innovation Center for Big Data and Brain Computing in Beihang University. His research interests include representation learning, machine learning and graph mining.
\end{IEEEbiography}
%\vskip -2.3\baselineskip

\begin{IEEEbiography}
[{\includegraphics[width=1in,height=1.25in,clip,keepaspectratio]{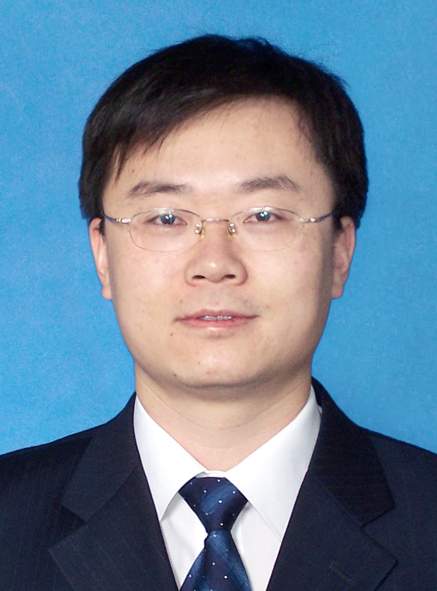}}]
{Jianxin Li} is currently a Professor with the State Key Laboratory of Software Development Environment, and Beijing Advanced Innovation Center for Big Data and Brain Computing in Beihang University. His current research interests include social network, machine learning, big data and trustworthy computing.
\end{IEEEbiography}
%\vskip +1\baselineskip

\begin{IEEEbiography}
[{\includegraphics[width=1in,height=1.25in,clip,keepaspectratio]{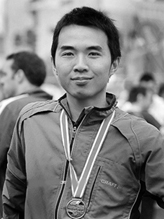}}] {Zheng Wang} is an Associate Professor with the
University of Leeds, UK. His research cuts across the boundaries of parallel program optimisation, systems security, and applied machine
learning.
\end{IEEEbiography}
%\vskip -1.6\baselineskip

\begin{IEEEbiography}[{\includegraphics[width=1in,height=1.25in,clip,keepaspectratio]{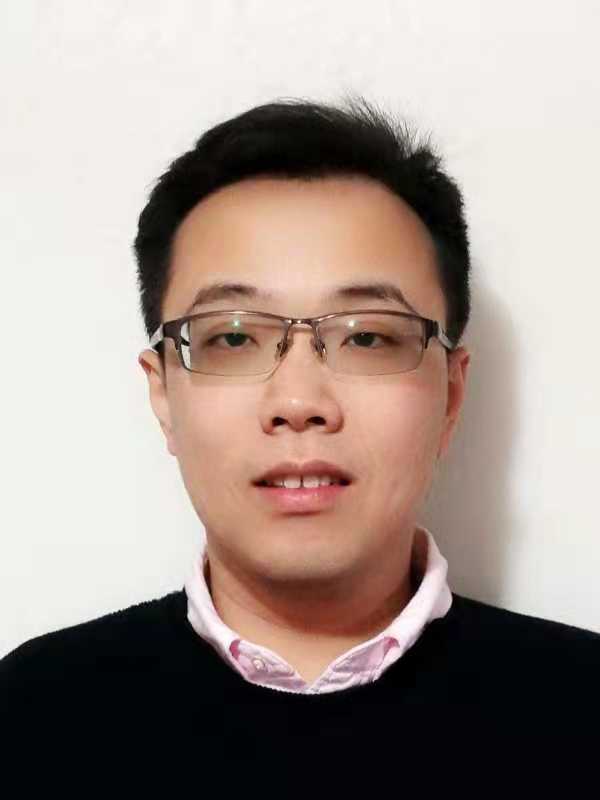}}] {Renyu Yang} is a Research Fellow with the University of Leeds, UK and adjunct researcher in Beijing Advanced Innovation Center for Big Data and Brain Computing in Beihang University. His research interests include reliable distributed systems, big data analytic at scale and applied machine learning.
\end{IEEEbiography}
%\vskip -1.6\baselineskip

\begin{IEEEbiography}[{\includegraphics[width=1in,height=1.25in,clip,keepaspectratio]{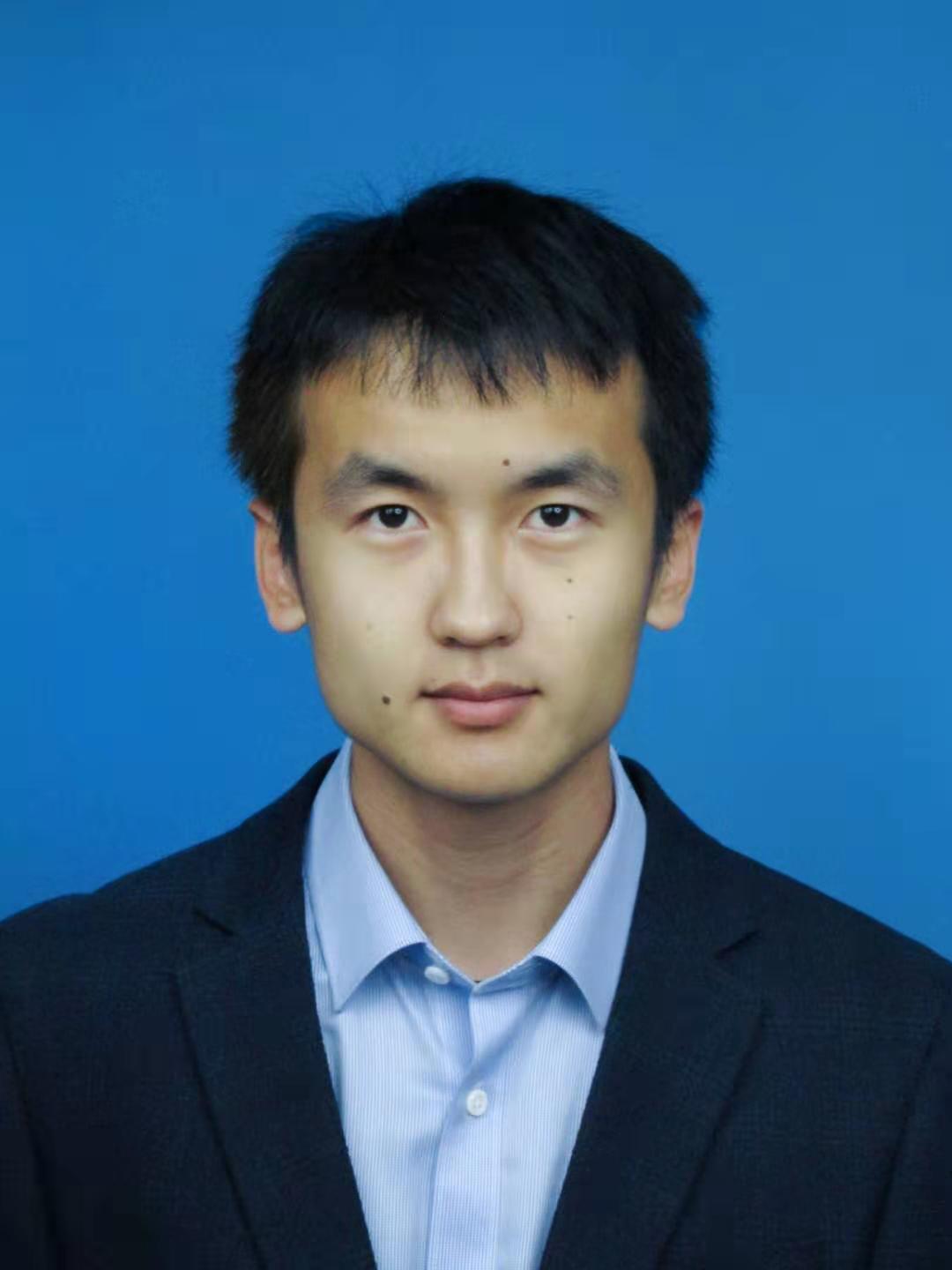}}] {Mingzhe Liu} is pursuing his MSc. at Beijing Advanced Innovation Center for Big Data and Brain Computing in Beihang University, Beijing, China. His research interests include urban computing and deep learning.
\end{IEEEbiography}
%\vskip -1.6\baselineskip

\begin{IEEEbiography}[{\includegraphics[width=1in,height=1.25in,clip,keepaspectratio]{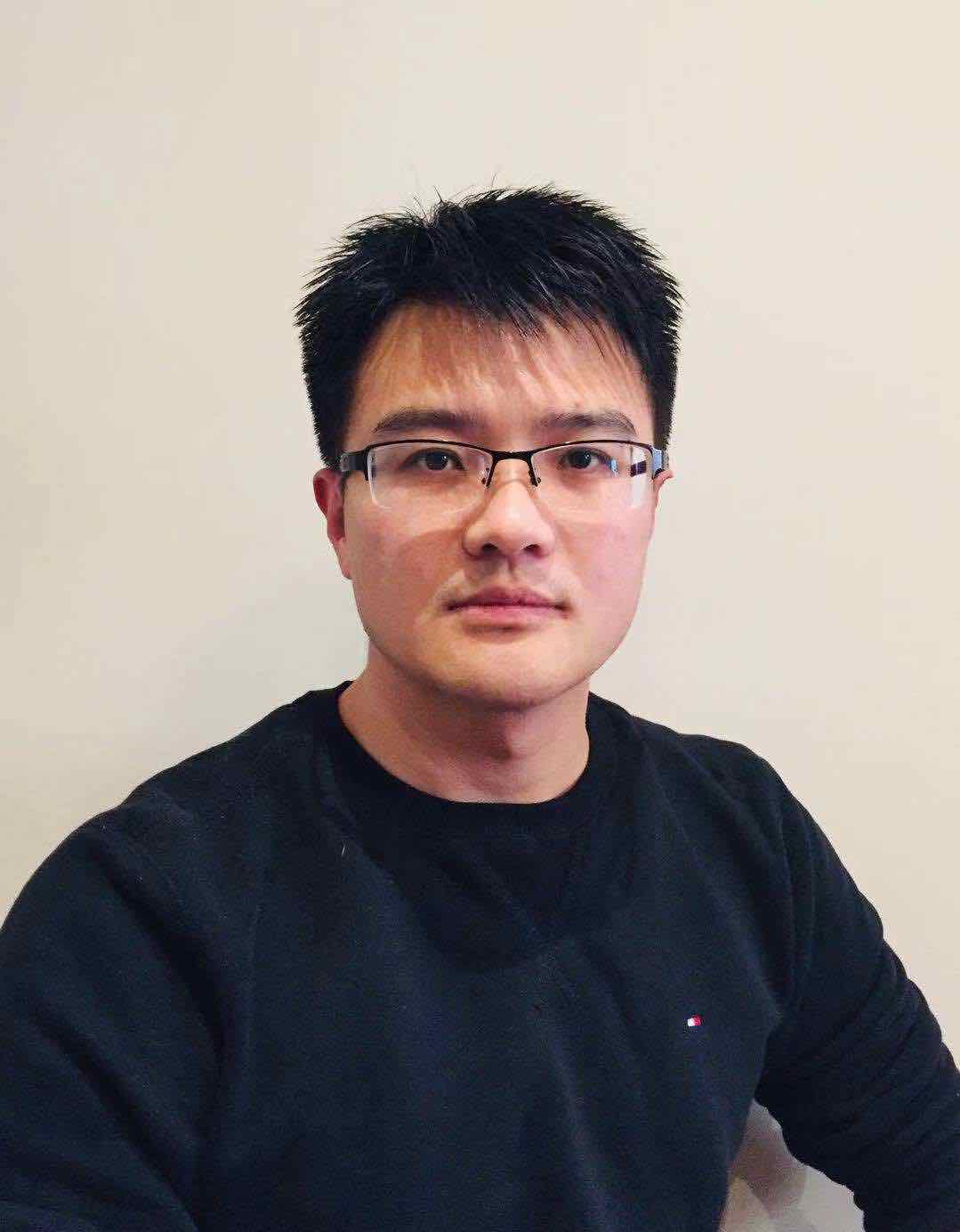}}] {Mingming Zhang} is a senior engineer in UrBrain Technology, Toronto, Canada. His research interests include urban computing and big data mining.
\end{IEEEbiography}
%\vskip -1.6\baselineskip

\begin{IEEEbiography}
[{\includegraphics[width=1in,height=1.25in,clip,keepaspectratio]{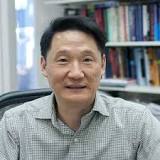}}]
{Philip S. Yu} is a Distinguished Professor and the Wexler Chair in Information Technology at the Department of Computer Science, University of Illinois at Chicago. Before joining UIC, he was at the IBM Watson Research Center, where he built a world-renowned data mining and database department. He is a Fellow of the ACM and IEEE. Dr. Yu was the Editor-in-Chiefs of ACM Transactions on Knowledge Discovery from Data (2011-2017) and IEEE Transactions on Knowledge and Data Engineering (2001-2004).
\end{IEEEbiography}
%\vskip -1.6\baselineskip

\begin{IEEEbiography}
[{\includegraphics[width=1in,height=1.25in,clip,keepaspectratio]{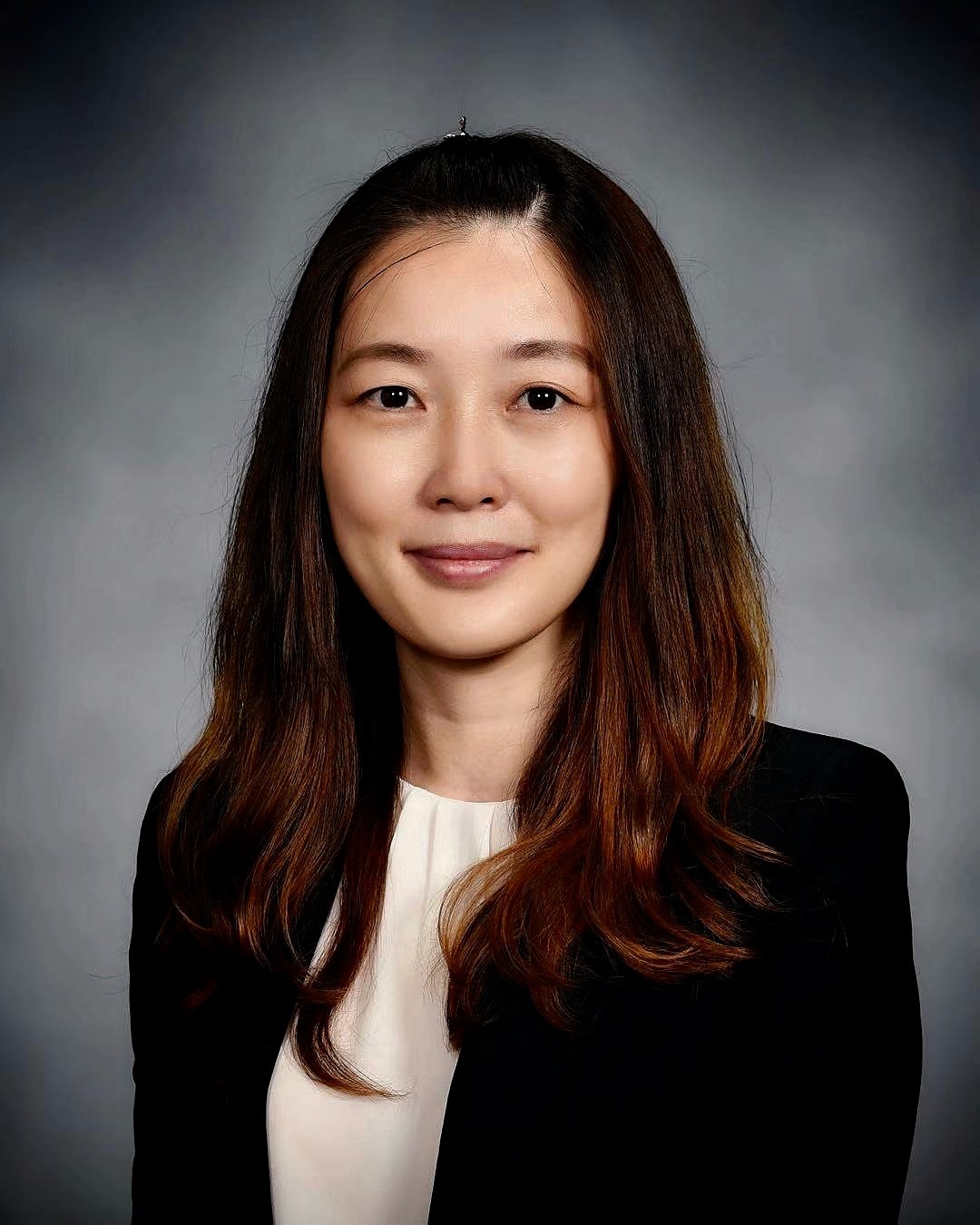}}]
{Lifang He} is currently an Assistant Professor in the Department of Computer Science and Engineering at Lehigh University. Before her current position, Dr. He worked as a postdoctoral researcher in the Department of Biostatistics and Epidemiology at University of Pennsylvania. Her current research interests include machine learning, data mining, tensor analysis, with major applications in biomedical data, neuroscience, or multimodal data.
\end{IEEEbiography}
%\vskip -1.6\baselineskip

\end{document}